\documentclass[sigconf]{acmart}

\AtBeginDocument{%
  }

\copyrightyear{2026}
\acmYear{2026}
\setcopyright{cc}
\setcctype{by}
\acmConference[KDD '26]{Proceedings of the 32nd ACM SIGKDD Conference on Knowledge Discovery and Data Mining V.1}{August 09--13, 2026}{Jeju Island, Republic of Korea}
\acmBooktitle{Proceedings of the 32nd ACM SIGKDD Conference on Knowledge Discovery and Data Mining V.1 (KDD '26), August 09--13, 2026, Jeju Island, Republic of Korea}
\acmPrice{}
\acmDOI{10.1145/3770854.3780195}
\acmISBN{979-8-4007-2258-5/2026/08}
\usepackage{array}
\usepackage{subcaption}
\usepackage{graphicx}      
\usepackage{color, soul}   
\usepackage{textcomp}      
\usepackage{url}           
\usepackage{verbatim}      
\usepackage{booktabs}      
\usepackage{tabularx}      
\usepackage{multirow}      
\usepackage{dsfont}        
\usepackage[ruled,vlined]{algorithm2e} 
\usepackage{comment}       
\usepackage{multicol}      

\usepackage{bm}

\newcommand{\RNum}[1]{\uppercase\expandafter{\romannumeral #1\relax}}
\newcommand{\Rnum}[1]{\lowercase\expandafter{\romannumeral #1\relax}}

\def\Appref#1{Appendix~\ref{#1}}

\def\Figref#1{Figure~\ref{#1}}

\def\Tabref#1{Table~\ref{#1}}

\def\eqref#1{equation~(\ref{#1})}
\def\Eqref#1{Equation~(\ref{#1})}

\def\0{\bm{0}} 
\def\1{\bm{1}}

\def\rvvarepsilon{{\boldsymbol{\varepsilon}}}

\def\rvc{{\mathbf{c}}}
\def\rvd{{\mathbf{d}}}

\def\rvf{{\mathbf{f}}}

\def\rvh{{\mathbf{h}}}

\def\rvm{{\mathbf{m}}}

\def\rvs{{\mathbf{s}}}

\def\rvw{{\mathbf{w}}}
\def\rvx{{\mathbf{x}}}
\def\rvy{{\mathbf{y}}}

\def\vx{{\bm{x}}}

\DeclareMathAlphabet{\mathsfit}{\encodingdefault}{\sfdefault}{m}{sl}
\SetMathAlphabet{\mathsfit}{bold}{\encodingdefault}{\sfdefault}{bx}{n}

\begin{document}

\title{TimeBridge: Better Diffusion Prior Design with Bridge Models for Time Series Generation}

\author{Jinseong Park}
\authornote{Equal contribution.}
\email{jinseong@kias.re.kr}
\affiliation{%
  \institution{Korea Institute for Advanced Study}
  \city{Seoul}
  \country{Republic of Korea}
}

\author{Seungyun Lee}
\authornotemark[1]
\email{rats96@snu.ac.kr}
\affiliation{%
  \institution{Seoul National University}
  \city{Seoul}
  \country{Republic of Korea}
}

\author{Woojin Jeong}
\email{jwj7955@snu.ac.kr}
\affiliation{%
  \institution{Seoul National University}
  \city{Seoul}
  \country{Republic of Korea}
}

\author{Yujin Choi}
\email{uznhigh@snu.ac.kr}
\affiliation{%
  \institution{Seoul National University}
  \city{Seoul}
  \country{Republic of Korea}
}

\author{Jaewook Lee}
\email{jaewook@snu.ac.kr}
\authornote{Corresponding author.}
\affiliation{%
  \institution{Seoul National University}
  \city{Seoul}
  \country{Republic of Korea}
}

\renewcommand{\shortauthors}{Jinseong Park et al.}

\begin{abstract}
Time series generation is widely used in real-world applications such as simulation, data augmentation, and hypothesis testing. Recently, diffusion models have emerged as the de facto approach to time series generation, enabling diverse synthesis scenarios. However, the fixed standard-Gaussian diffusion prior may be ill-suited for time series data, which exhibit properties such as temporal order and fixed time points.
In this paper, we propose \textit{TimeBridge}, a framework that flexibly synthesizes time series data by using diffusion bridges to learn paths between a chosen prior and the data distribution. We then explore several prior designs tailored to time series synthesis. Our framework covers (i) data- and time-dependent priors for unconditional generation and (ii) scale-preserving priors for conditional generation. Experiments show that our framework with data-driven priors outperforms standard diffusion models on time series generation.
\end{abstract}




\maketitle
\section{Introduction}

Time series generations have expanded their applications by serving roles in data augmentation \cite{wen2021time}, data privacy \cite{wang2024igamt}, and hypothesis‐driven simulations of data streams \cite{xia2024market}. Even imputation and forecasting can be treated as special cases of time series generation.
With advances in deep generative modeling, practitioners have developed VAE‐based \cite{naiman2024generative}, GAN‐based \cite{yoon2019time}, and diffusion‐based methods \cite{ho2020denoising,karras2022elucidating}.
Diffusion approaches have shown strong performance, yielding high‐fidelity and diverse samples \cite{tashiro2021csdi,lopez2023diffusionbased} by estimating the reverse trajectory of a forward noising process toward a standard Gaussian prior \cite{song2021scorebased}.


Nevertheless, controlling the diffusion process remains difficult \cite{harvey2022flexible,du2023flexible} when incorporating domain-specific constraints or time-dependent properties in time series. For instance, \citet{coletta2024constrained} highlighted that generating data under specific bull or bear market patterns requires tailored embedding or penalty functions for sampling guidance. As standard diffusion forces a standard Gaussian for prior distribution, it is hard to control the diffusion model to reflect the important properties of time series, such as sequential order, historical statistics, or temporal dependencies \cite{kim2022reversible, zeng2023transformers}, during the whole reverse trajectories.

\begin{figure}[!t] 
\centering
\includegraphics[width=0.98\linewidth]{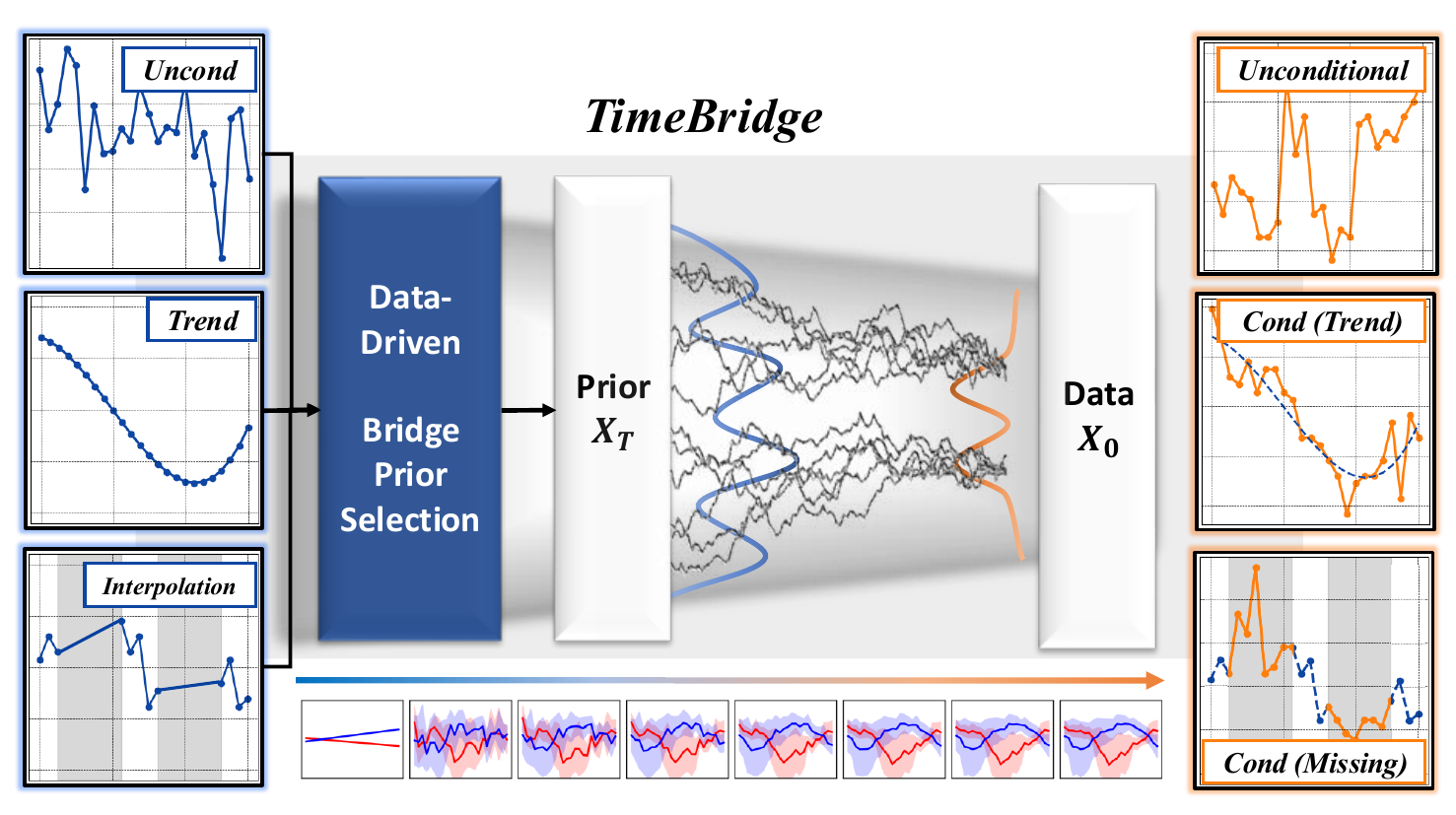}
 \caption{Illustration of the proposed time series diffusion bridge framework enabling various prior selections.}
 \label{fig:method}
 \vspace{-0.6cm}
\end{figure}

\begin{table*}[!ht]
\centering
\caption{Capability comparison of time series generation. $\bigcirc$ = supported, $\triangle$ = partially supported, $\times$ = not supported.}
\label{tab:compar}
\resizebox{0.9\textwidth}{!}{%
\begin{tabular}{l|cc|cccccc}
\toprule
Model & Sampler & Steps & Uncond.\ Gen. & Constrained Gen. & Decomp.~Layer & Bridge & Flexible Prior \\
\midrule
SSSD \cite{lopez2023diffusionbased}       & DDPM      & 1000          &  $\bigcirc$     & $\triangle$ & $\times$ & $\times$ & $\times$ \\
DiffTime \cite{coletta2024constrained}    & DDPM      & 1000          & $\triangle$    & $\bigcirc$  & $\times$ & $\times$ & $\times$ \\
Diffusion-TS \cite{yuan2024diffusionts}   & Langevin  & 500, 1000     & $\bigcirc$  & $\triangle$ & $\bigcirc$& $\times$ & $\times$ \\
\textbf{TimeBridge}                       & DDBM \cite{zhou2024denoising} & 119 ($\downarrow$) & $\bigcirc$  & $\bigcirc$ & $\bigcirc$ & $\bigcirc$ & $\bigcirc$ \\
\bottomrule
\end{tabular}
}
 \vspace{-0.25cm}
\end{table*}


When viewing a generative model as optimal transport between data and prior distributions, the choice of prior distribution significantly impacts the training difficulty \cite{kollovieh2024flow}. Rather than mapping data to a standard normal prior, recent studies altered the diffusion process with data-dependent forward and reverse processes for controlled generation \cite{han2022card,yu2024constructing} or developed diffusion steps aligned with actual timestamps in time series domains \cite{bilovs2023modeling,chen2023provably}.

In this paper, we extend this line of work by leveraging diffusion bridges \cite{Schrodinger1932theorie,de2021diffusion,zhou2024denoising} to control prior distributions in time series diffusion models, enabling optimal transport between data and carefully chosen priors. While diffusion bridges have yielded impressive results in conditional generation settings such as image-to-image translation \cite{zhou2024denoising} and text-to-speech \cite{chen2023tts}, existing work has focused primarily on how to connect given data and prior distributions, with limited attention to the crucial question of prior selection.

\begin{figure}[!t] 
\centering
\includegraphics[width=\linewidth]{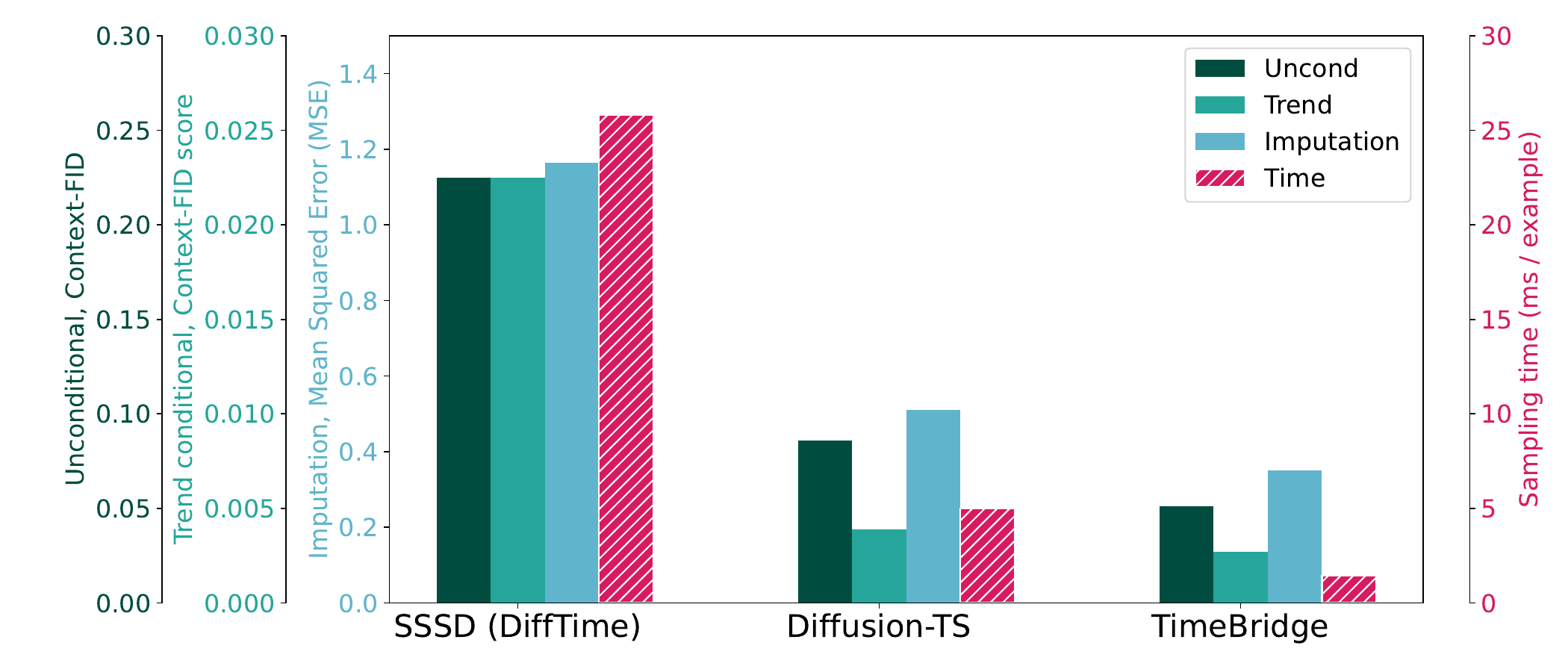}
 \caption{Performance measures (lower is better) on different tasks and sampling time (lower is better) of time series diffusion models. The proposed TimeBridge achieves both generalization in various tasks and sampling efficiency.}
 \label{fig:acc_time}
 \vspace{-0.3cm}
\end{figure}

To address this gap in time series generation, we propose \textit{TimeBridge}, a diffusion bridge model for time series (illustrated in \Figref{fig:method}) with an architecture tailored to sequential data based on Diffusion-TS \cite{yuan2024diffusionts}. We then investigate which priors better replace the standard Gaussian prior, enabling more efficient synthesis and higher-quality results across diverse settings. 
Compared to standard diffusion baselines summarized in \Tabref{tab:compar}, TimeBridge shows higher accuracy in both unconditional and conditional tasks across various datasets, as demonstrated in \Figref{fig:acc_time}. 

Our key contributions with flexible prior selection are:
\begin{itemize}
    \item 
    We explore diverse priors to alter the standard Gaussian for capturing time series characteristics and propose a new diffusion bridge model tailored to time series generation.
    \item For unconditional generation, we design data- and time-dependent priors that reflect intrinsic temporal properties.
    \item For conditional generation, we embed scale constraints into a pairwise prior and propose \textit{point-preserving sampling}, extending its application to imputation tasks.
\end{itemize}
\noindent
We publish the code at {\color{blue}
\url{https://github.com/JinseongP/TimeBridge}.}

\section{Preliminaries}
Consider the multivariate time series $\rvx$ defined within the sample space $\mathcal{X} = \mathbb{R}^{d \times \tau}$, with the dimensionality $d$ and the time length $\tau$. For each index $i \in \{1, \ldots, N\}$ among the $N$ samples, each time series is denoted as $\rvx^i = (\vx^i(1), \ldots, \vx^{i}(\tau)) \in \mathcal{X}$, given the data $\vx^{i}(k)\in\mathbb{R}^d$ for each timestamp $k\in\{1,\ldots,\tau\}$. For diffusion steps $t\in\{0,\ldots,T\}$, we denote intermediate  time series samples as $\rvx_t$, and for a specific instance $i$ as $\rvx_t^i$.

\noindent
\paragraph{\textbf{Diffusion Models. }}
In standard diffusion models \cite{ho2020denoising}, the diffusion process is constructed by gradually injecting noise into samples $\rvx_0$ drawn from the data distribution $p_0$, forwarding them into a fixed standard Gaussian distribution $p_T = \mathcal{N}(\boldsymbol{0},\boldsymbol{I})$. The corresponding forward SDE is: 
\begin{equation}
\label{eqn:forward}
    d\rvx_t = \rvf(\rvx_t,t)dt + g(t)d\rvw_t,
\end{equation}
where $\rvf:\mathbb{R}^d \times [0,T] \rightarrow \mathbb{R}^d$ represents the drift function, $g:[0,T] \rightarrow \mathbb{R}$ denotes the diffusion coefficient, and $\rvw_t$ is the Wiener process. As diffusion models learn to reverse the \Eqref{eqn:forward}, the reverse SDE  \cite{anderson1982reverse} is formulated as follows:
\begin{equation}
    d\rvx_t = [\rvf(\rvx_t,t) - g(t)^2\nabla_{\rvx_t} \text{log}p(\rvx_t)]dt + g(t)d\bar{\rvw}_t,
\end{equation}
where $p(\rvx_t)$ refers to the probability density of $\rvx_t$. Diffusion models learn to match the score function $\nabla_{\rvx_t} \text{log}p(\rvx_t)$ for denoising \cite{song2021scorebased}.

\paragraph{\textbf{Optimal Transport and Schrödinger Bridge. }}
The aforementioned standard diffusion models assume a standard Gaussian in the prior, and thus cannot be extended to general prior distributions. On the other hand, the optimal transport view for generative modeling is to find a trajectory of two arbitrary distributions. Specifically, the Schrödinger Bridge (SB) provides an entropy-regularized solution by minimizing the KL divergence between path measures \citep{chen2021likelihood}:

\begin{equation}
    \underset{p \in \mathcal{P}_{[0,T]}}{min}D_{KL}(p \lVert p^{ref}), \quad s.t. \ p_0 = p_{data}, \ p_T = p_{prior},
\end{equation}
where, $\mathcal{P}_{[0,T]}$ denotes the space of path measures on the interval $[0,T]$, $p^{\text{ref}}$ is a reference path measure, and $p_0,p_T$ are the marginals of $p$. By choosing a forward stochastic differential equation (SDE) associated with the reference process $p^{\text{ref}}$, the SB can be reformulated by the following forward-backward SDEs:
\begin{equation}
    \begin{split}
     d\rvx_t = [\rvf(\rvx_t,t)+g^2(t) \nabla \log \boldsymbol{\Psi}_t(\rvx_t)]dt + g(t) d\rvw_t, \ \rvx_0 \sim p_{data}, \\
    d\rvx_t = [\rvf(\rvx_t,t)-g^2(t) \nabla \log \hat{\boldsymbol{\Psi}}_t(\rvx_t)]dt + g(t) d\bar{\rvw}_t,  \ \rvx_T \sim p_{prior},
    \end{split}
\end{equation}
where $ \nabla \log \boldsymbol{\Psi}_t (x_t) $ and $ \nabla \log \hat{\boldsymbol{\Psi}}_t (\rvx_t) $ are corresponding partial differential equations.  In earlier practice, the Schrödinger bridge-based trajectory simulation typically relied on iterative methods \citep{kullback1968probability}, making it less widely used than diffusion models due to computational complexity and training stability \cite{zhou2024denoising}. The details of Schrödinger Bridge, including iterative solution, are explained in Appendix \ref{app:diffusion_bridge}.

\paragraph{\textbf{Diffusion Bridge. }}
Recently, a non-iterative approach to diffusion bridges has emerged, which constructs diffusion processes towards any endpoint by applying Doob's h-transform \cite{doob1984classical} to \Eqref{eqn:forward}. With a fixed endpoint $\rvy$, the objective is to trace the trajectory of the transformed path for the diffusion bridge as follows:

\begin{align}
\begin{split}
    d\rvx_t = [\rvf(\rvx_t,t) + g(t)^2\rvh(\rvx_t,t,\rvy,T)]dt + g(t)d\rvw_t, \\
    \rvx_0 \sim q_{data}(\rvx), \rvx_T=\rvy,
    \label{eq:bridge_forward}
\end{split}
\end{align}
where $\rvh(\rvx,t,\rvy,T) = \nabla_{\rvx_t} \text{log}p(\rvx_T|\rvx_t)|_{\rvx_t=\rvx,\rvx_T=\rvy}$.
For a given data distribution $q_{data}(\rvx,\rvy)$, the diffusion bridge \cite{Schrodinger1932theorie, de2021diffusion} aims to transport $\rvx_0$ to $\rvx_T$ where $(\rvx_0,\rvx_T) = (\rvx,\rvy) \sim q_{data}(\rvx,\rvy)$ while following above forward process. Similar to the standard diffusion model, we can construct the reverse SDE as follows:
\begin{align}
\begin{split}
    d\rvx_t = [\rvf(\rvx_t,t) - g(t)^2(\rvs(\rvx_t,t,\rvy,T)- \\ \rvh(\rvx_t,t,\rvy,T))]dt
     + g(t)d\bar{\rvw}_t,
    \label{eq:bridge_reverse}
\end{split}
\end{align}
where $\rvx_T=\rvy$ and its score function is calculated as
\begin{equation}
    \rvs(\rvx,t,\rvy,T) = \nabla_{\rvx_t} \text{log}q(\rvx_t|\rvx_T)|_{\rvx_t=\rvx,\rvx_T=\rvy}.
    \label{eq:score}
\end{equation}
For training, diffusion bridge can leverage score-based methods by learning the score function $\rvs(\rvx,t,\rvy, T)$.


Diffusion bridge models have been investigated due to their ability to flexibly map in various domains, including image-to-image translation \cite{li2023bbdm, zhou2024denoising} and text-to-speech synthesis \cite{chen2023tts}, compared to standard diffusion models. Recently, adapting the advances of standard diffusion, such as consistency \cite{he2024consistency}, and deterministic samplers \cite{zheng2025diffusion}, has been further investigated in the diffusion bridge. 

Note that although SB theory covers the concepts of diffusion bridges, we separate SB methods with iterative fitting and score-based diffusion bridge methods for clarity in this paper. 
\section{Framework}
\subsection{Scenarios}
For time series generation, we cover both unconditional and conditional generations.
\textbf{Unconditional generation} considers the total dataset containing $N$ samples, denoted as $\mathcal{D} = \{{\rvx^i}\}_{i=1}^N$. Our goal is to train a deep generative model to mimic the input dataset and yield synthetic samples statistically similar to a given time series dataset. We aim to make the distribution $p(\hat \rvx|\cdot)$ of synthetic data $\hat \rvx$, similar to the input data distribution $p(\rvx)$.

\textbf{Conditional generation} further utilizes the paired condition $\rvy$  to guide time series generation of each data sample $\rvx$ as $\mathcal{D}=\{(\rvx^i,\rvy^i)\}_{i=1}^{N}$. Thus, we try to make the conditional distribution  $p(\hat \rvx|\rvy)$ of synthetic data $\hat \rvx$ closely mirror the conditional distribution of input data $p(\rvx|\rvy)$. Considering that the condition in time series is often also a time series as $\rvy \in \mathcal{X}$, \citet{coletta2024constrained} divided the constraints into two types: (i) \textbf{soft constraint} to guide the time series generation such as trend, or (ii) \textbf{hard constraint} where we should follow during synthesis such as fixed points.
In a broader view, the unmasked part in imputation or the historical data in forecasting can be considered as conditions for generation. 

\begin{figure*}[!t]
    \centering
    \begin{subfigure}{0.16\textwidth}
        \includegraphics[width=\linewidth]{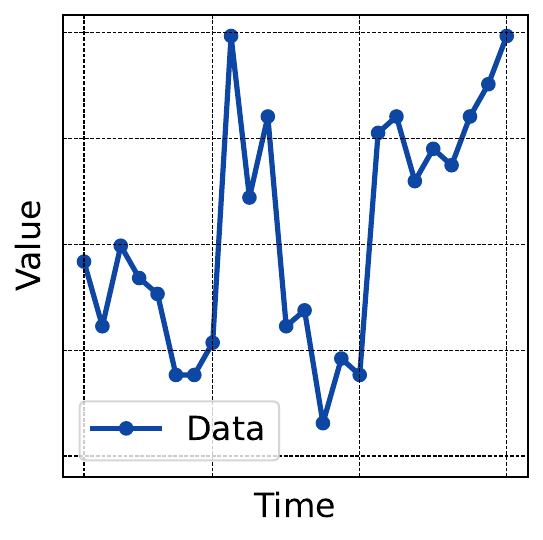}
        \caption{Unconditional}
        \label{fig:toy_uncond}
    \end{subfigure}
    \begin{subfigure}{0.16\textwidth}
        \includegraphics[width=\linewidth]{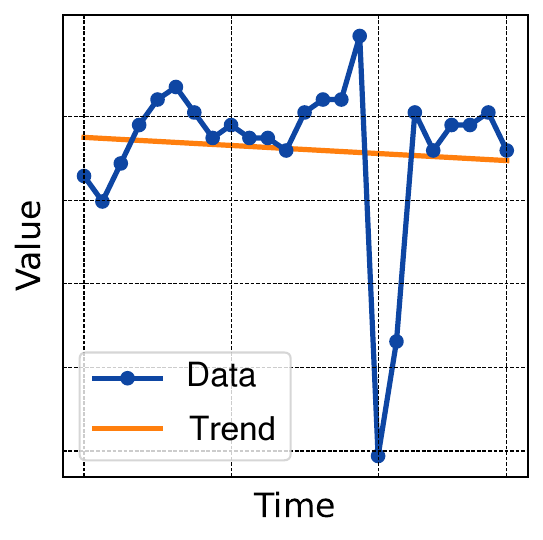}
        \caption{Linear trend}
        \label{fig:toy_trend_linear}
    \end{subfigure}
    \begin{subfigure}{0.16\textwidth}
        \includegraphics[width=\linewidth]{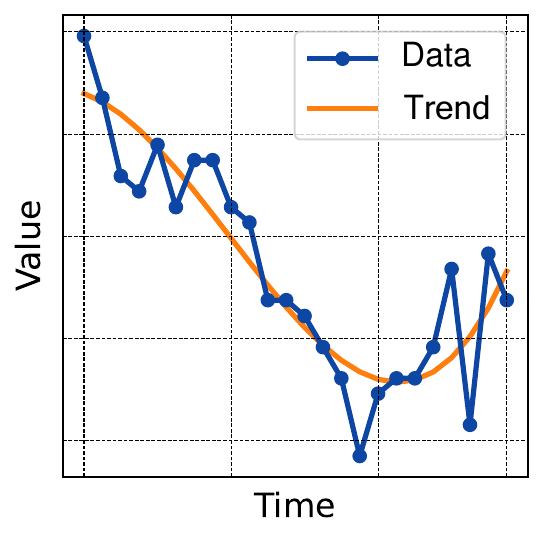}
        \caption{Polynomial trend}
        \label{fig:toy_trend_poly}
    \end{subfigure}
    \begin{subfigure}{0.16\textwidth}
        \includegraphics[width=\linewidth]{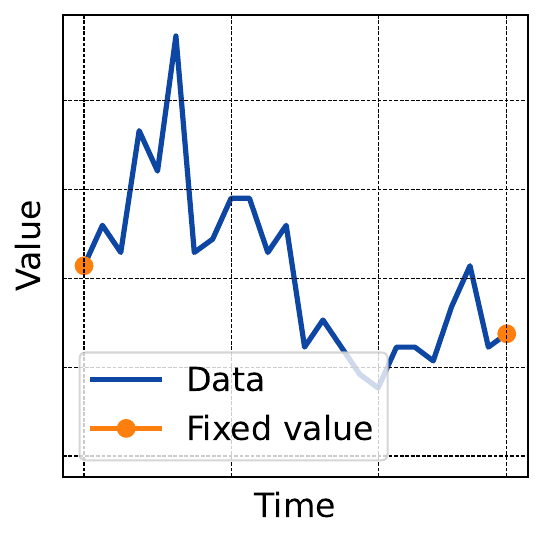}
        \caption{Fixed point}
        \label{fig:toy_fixed}
    \end{subfigure}
    \begin{subfigure}{0.16\textwidth}
        \includegraphics[width=\linewidth]{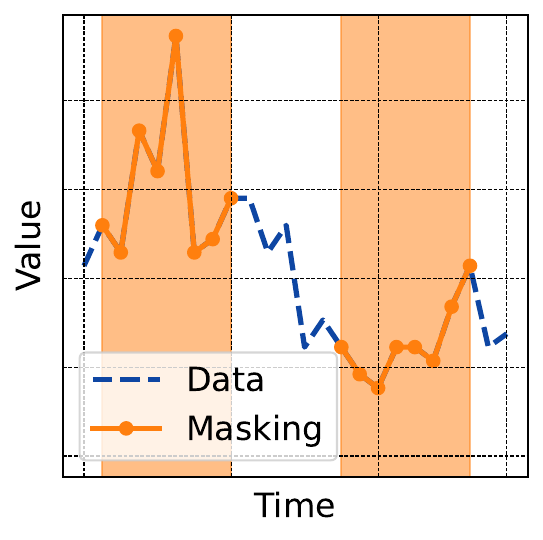}
        \caption{Missing value}
        \label{fig:toy_masking}
    \end{subfigure}
    \vspace{-0.2cm}
    \caption{Illustration of time series generation situations under various conditions.}
    \label{fig:situation22}
\vspace{-0.3cm}
\end{figure*}

We illustrate time series generation situations on various conditions, including unconditional, linear trend, polynomial trend, fixed point, and missing value, in \Figref{fig:situation22}.

\subsection{Diffusion Bridge for Time Series}
To leverage the prior selections, we propose \textbf{TimeBridge}, a diffusion bridge tailored to time‐series synthesis. Our approach produces high-quality samples with fewer sampling steps and allows flexible prior selection.  We now provide a detailed description of the sampler designs, network architecture, and overall framework. 

\paragraph{\textbf{Diffusion Bridge Sampler}}
To approximate the score $\rvs(\rvx,t,\rvy,T)$ in \Eqref{eq:score}, we adopt the Denoising Diffusion Bridge Model (DDBM) sampler \cite{zhou2024denoising}. Unlike prior time‐series methods based on the Schrödinger bridge \cite{chen2023tts,garg2024soft}, DDBM aligns with recent diffusion frameworks and enables efficient bridge sampling. Accordingly, we exploit the tractable marginal distribution of $\rvx_t$ by setting $\rvx_t = \alpha_t \rvx_0 + \sigma_t \rvvarepsilon$, where $\alpha_t$ and $\sigma_t$ are noise‐schedule functions and $\rvvarepsilon \sim \mathcal{N}(\boldsymbol{0},\boldsymbol{I})$. Under the variance‐preserving (VP) schedule, the marginal distribution at time $t$ becomes
\begin{align}
&q(\rvx_t|\rvx_0, \rvx_T) = \mathcal{N}\left( \hat{\boldsymbol\mu}_t,\hat{\sigma}^2_t\mathbf{I}\right), \label{eq:rvx_t}\\
&\hat{\boldsymbol{\mu}}_t = \textstyle \frac{SNR_T}{SNR_t} \frac{\alpha_t}{\alpha_T}\rvx_T + \alpha_t \rvx_0\left(1-\frac{SNR_T}{SNR_t}\right),\\
&\hat{\sigma}^2_t = \textstyle \sigma^2_t\left(1-\frac{SNR_T}{SNR_t}\right),
\end{align}
where signal-to-ratio (SNR) is defined as $SNR_t = {\alpha^2_t}/{\sigma^2_t}$.
Since the mean $\hat{\boldsymbol\mu}_t$ is a linear combination of $\rvx_0$ and $\rvx_T$ at time $t$, the data scale is preserved for translation. 

For score matching, \citet{karras2022elucidating} proved that if a denoiser function $D(\rvx;\sigma)$ minimizes the expected error for samples drawn from $p_{data}$, then $D(\rvx;\sigma)$ becomes the linear function of score function, i.e., if $D(\rvx; \sigma)$ minimizes $ \mathbb{E}_{\rvy \sim p_{data}} \mathbb{E}_{\boldsymbol{\rvvarepsilon} \sim \mathcal{N}(\mathbf{0}, \sigma^2 \boldsymbol{I})} \lVert D( \rvy + \boldsymbol{\rvvarepsilon}; \sigma) -\rvy \rVert^2_2 $, then $\nabla_\rvx \text{log}p(x; \sigma) = (D(\rvx; \sigma) - \rvx)/ \sigma^2 $. Thus, we match $D_{\theta}$ directly, which is equivalent to training the score function \cite{karras2022elucidating} as
\begin{equation}
\label{eq:reparam}
\begin{aligned}
  &\nabla_{\rvx_t} \text{log}q(\rvx_t|\rvx_T) \approx \rvs({D_\theta}, \rvx_t, t, \rvx_T, T) \\ &:= \frac{\rvx_t - (\frac{SNR_T}{SNR_t}\frac{\alpha_t}{\alpha_T}\rvx_T + \alpha_t D_{\theta}(\rvx_t,t, \rvx_T)(1-\frac{SNR_T}{SNR_t}))}{\sigma^2_t(1-\frac{SNR_T}{SNR_t})},
\end{aligned}
\end{equation}
where $D_{\theta}(\rvx_t,t,\rvx_T)$ is the model output as a denoiser.

\paragraph{\textbf{Architecture Design}} 
To obtain $D_{\theta}(\rvx_t,t,\rvx_T)$, we follow the backbone of Diffusion-TS \cite{yuan2024diffusionts}, which employs an encoder–decoder transformer.
Let $w_{(\cdot)}^{i,t}$ denote the input of the interpretable layers for index $i\in{1,\ldots,K}$ at diffusion step $t$.
Each decoder block is split into {trend} and {seasonal} synthesis layers.
First, the trend component is synthesized with a polynomial regressor:

\begin{equation}
    V_{tr}^t = \Sigma_{i=1}^K(C \cdot \text{Linear}(w_{tr}^{i,t}) + X_{tr}^{i,t}),
\end{equation}
where $X_{tr}^{i,t}$ is the mean value of the output of the $i^{th}$ decoder block with the tensor multiplication $(\cdot)$ to $C$.

For seasonal components, we use the Fourier transformation as
 \begin{multline}
         S_{i,t} = \Sigma_{k=1}^K A_{i,t}^{\kappa_{i,t}^{(k)}} [cos(2\pi f_{\kappa_{i,t}^{(k)}}\tau c + \Phi_{i,t}^{\kappa_{i,t}^{(k)}}) + cos(2\pi \bar{f}_{\kappa_{i,t}^{(k)}}\tau c + \bar{\Phi}_{i,t}^{\kappa_{i,t}^{(k)}})],
 \end{multline}
 where $\text{argTopK}$ is to get the top $K$ amplitudes and $f_k$ represents the Fourier frequency of the index $k$ and $\bar{(\cdot)}$ denotes $(\cdot)$ of the conjugates. $ A_{i,t}^{(k)}$ and $\Phi_{i,t}^{(k)}$ are the phase and amplitude of the $k$-th frequency, respectively, after the Fourier transform $\mathcal{F}$ as
\begin{equation}
    A_{i,t}^{(k)} = \lvert \mathcal{F}(w_{seas}^{i,t})_k \rvert , \Phi_{i,t}^{(k)} = \phi(\mathcal{F}(w_{seas}^{i,t})_k),
\end{equation}
\begin{equation}
    \kappa_{i,t}^{(1)},...,\kappa_{i,t}^{(K)} = \underset{k \in \{ 1,..., \lfloor \tau/2 \rfloor +1 \}}{\text{argTopK}} \{ A_{i,t}^{(k)}\}.
\end{equation}

\paragraph{\textbf{Overall Framework}}
Based on the previous subsections, we can design the output of the denoiser $D_\theta$ with both conditioning on noisy data $\rvx_t$ dependent on data $\rvx_0$ and prior $\rvx_T$ as follows:  
\begin{equation} \label{eq:d_theta}
D_\theta = V_t^{tr}(\theta, \rvx_t, t, \rvx_T) + \sum_{i=1}^K S_{i,t}(\theta, \rvx_t, t,\rvx_T) + R(\theta, \rvx_t, t, \rvx_T), 
\end{equation}
where $V_t^{tr}$ is the output for the trend synthesis layer, $S_{i,t}$ is the output for each seasonal synthesis layer $i$ among $K$ layers and $R$ is the output for estimated residual. Note that calculating \Eqref{eq:d_theta} is unique in diffusion bridge since the standard diffusion models \cite{yuan2024diffusionts} do not consider the dependency on prior $\rvx_T$ for calculating $\rvx_t$.

For training, our objective function $\mathcal{L}_\theta$ is formulated as follows:
\begin{equation}
\label{eq:loss}
\begin{aligned}
    \mathcal{L}_{\theta} = \mathbb{E}_{t, \rvx_0} [ w_t ( \lVert \rvx_0 - D_\theta(\rvx_t,t, \rvx_T) \rVert^2 \\  + \lambda \lVert FFT(\rvx_0) - FFT(D_\theta(\rvx_t,t, \rvx_T)) \rVert^2 )].
\end{aligned}
\end{equation}
$w_t$ indicates the weight scheduler of the loss function, and $\lambda$ indicates the strength of the Fourier transform. 

As time series data possess stochasticity \cite{shen2023non}, we directly predict 
$D_\theta$ rather than using the reparametrization technique from noise prediction used in other domains \cite{karras2022elucidating,zhou2024denoising}. 
Moreover, we utilize the second-order Heun sampler \cite{ascher1998computer, karras2022elucidating}, achieving both quality and efficiency over first-order methods.

Overall, we name the proposed framework as \textbf{TimeBridge}, a flexible time series generation method that enables the selection of data-driven priors tailored to time series data based on (i) a flexible diffusion framework with diffusion bridge samplers and (ii) an interpretable decomposition-based time series structure.


\section{Prior Design for Time Series Synthesis}
\label{sec:prior}

Based on the flexibility of choosing priors with the proposed TimeBridge, we hypothesize that the standard Gaussian prior might not be the optimal choice. Thus, we investigate a general time series synthesis framework by analyzing the prior selection in diffusion models based on the following \textbf{research questions (RQs)}:

\textbf{RQ1}. \textit{Are data-dependent priors effective for time series?}

\textbf{RQ2}. \textit{How to model temporal dynamics with priors?}

\textbf{RQ3}. \textit{Can constraints be used as priors for diffusion?}

\textbf{RQ4}. \textit{How to preserve data points with diffusion bridge?}

\subsection{Data- and Time-dependent Priors for Time Series}
For unconditional tasks, we examine the benefit of better priors depending on data and temporal properties of time series.

\paragraph{\textbf{Data-Dependent Prior}}
To investigate \textbf{RQ1}, we suggest using data prior to having a data-dependent distribution to better approximate the data distribution. Sampling from $\mathcal N (\mathbf{0},\mathbf{I})$ lacks data-specific information, which can be enhanced by enforcing that the priors capture data scale and temporal dependencies. Recently, the usages of data-dependent prior have been investigated for audio \cite{popov2021grad,lee2022priorgrad} and image \cite{yu2024constructing} domains. Therefore, we test the use of data-dependent prior for time series by setting the prior distribution as follows:
\begin{equation}
    \label{eq:mean}
    \rvx_T\sim\mathcal N (\boldsymbol{\mu},\texttt{diag}(\boldsymbol\sigma^2)),
\end{equation}
where $\boldsymbol{\mu}$ and $\boldsymbol\sigma^2$ are the mean and variance of the data samples for each feature and timestamp independently calculated in the same dimension of the time series (e.g., for ETTh, resulting in $\boldsymbol{\mu}$ with a shape of $(24 \times 7)$). $\texttt{diag}(\boldsymbol\sigma^2)$ indicates the diagonal covariance matrix of each element. 

\paragraph{\textbf{Time-Dependent Prior}}

In response to \textbf{RQ2}, we focus on capturing temporal dependencies in noise addition towards prior distribution.
Recently, \citet{bilovs2023modeling} injected stochastic processes instead of random noise to preserve the continuity of temporal data and  
\citet{han2022card} utilized the pre-trained model for prediction and set the mean of prior as the output for regression.

As a solution, we employ Gaussian processes (GPs) for prior selection, which is effective for modeling temporal data \citep{bilovs2023modeling, ansari2024chronos}. GP is a distribution over function that defines the joint variability of an arbitrary pair of time points. For given sequence of timestamps $ \Omega = \{1,\ldots,\tau\}$ and a function $\boldsymbol{f}_{GP}$ of timestamps that follows a GP with mean function $\boldsymbol{m}(\cdot)$ and the covariance function $\boldsymbol{\Sigma}(\cdot)$, we can draw $\rvx_T$ as follows:
\begin{equation}
\label{eq:gp_prior}
\boldsymbol{f}_{GP} \sim {GP}(\boldsymbol{m}, \boldsymbol{\Sigma}), \ \   \rvx_T \sim \boldsymbol{f}_{GP}(\Omega).
\end{equation}
In this paper, we construct the time-dependent prior by aligning the GP with the aforementioned data-dependent prior at given time points. Specifically, we add the radial basis function (RBF) kernel $\mathcal{K}(i,j) = \eta exp(-\gamma|i - j|^2)$ for timestamps $i$ and $j$ into \Eqref{eq:mean}. Details for sampling method are in \Appref{app:add_exp}.

\subsection{Scale-preserving for Conditional Priors}
We now consider the condition $\rvy \in \mathcal{X}$, provided as a time series, which is the most prevalent case for real-world settings. 
We explore using a prior on the same data scale instead of relying solely on conditional embeddings in standard diffusion.

\paragraph{\textbf{Soft constraints with given trends}}
To answer \textbf{RQ3}, we argue that the diffusion bridge is well-suited for preserving a given condition by setting the prior as the condition. As one of the common conditional setups, we consider the task of generating data that follows a given trend.


Conditional generation based on trend is straightforward by (i) setting the pair-wise trend to prior $\rvx_T=\rvy$ and (ii) training the translation from trend to data. As the expected mean of $\rvx_t$ is a linear combination of $\rvx_0$ and $\rvx_T=\rvy$ in \Eqref{eq:rvx_t}, the model learns the correction starting from the trend samples, utilizing the same data scale. In contrast, standard diffusion models are unsuitable for translating conditions into synthetic data; instead, they generate new data based on a specific trend condition, typically requiring additional steps such as providing guidance or correcting intermediate data samples $\rvx_t$ during sampling. Thus, we can also eliminate the need for penalty functions or corrections for sampling.

\paragraph{\textbf{Hard constraints with fixed points}}

In \textbf{RQ4}, preserving data points of hard constraints can be important in conditional tasks. To ensure the fixed point condition during sampling, we introduce a novel \textit{point-preserving sampling} method that prevents adding noise to the values to be fixed. This is a unique trait of our framework because standard diffusion models cannot preserve identity trajectories to the constrained values from Gaussian noise. 

The direct case of fixed point constraints is imputation. With the mask $\rvm$ indicating the missing value as 0, we construct the condition $\rvc=\rvy\odot\rvm$ for imputation where $\odot$ indicates the Hadamard product. However, unlike soft constraints that apply to all values in $\rvy \in \mathbb{R}^{\tau \times d}$, the condition for imputation $\rvc \in \mathbb{R}^{\tau \times d}$ has missing values, which should be interpolated to data-dependent values rather than zeros. Therefore, we build local linear models for condition-to-time series imputation. We adopt \textit{linear spline interpolation} \cite{de1978practical} to generate a prior for fixed point constraints. In a feature-wise manner, let us consider $x\in\mathbb{R}$ as a single-dimensional input of $\vx\in\mathbb{R}^d$. With a set $K$ of feature-wise observed times for $x$, the prior $x_T$ with the interpolation of each feature with timestamp $k$ is:
\begin{equation}
    \label{eq:interpolate}
    \begin{aligned}
        x_T(k) = x(k_{j-1}) + \frac{x(k_{j}) - x(k_{j-1})}{k_{j} - k_{j-1}} (k-k_{j-1}), \\ \quad k_{j-1} \leq k < k_{j},\quad j = 1, \ldots, |K|-1.
    \end{aligned}
\end{equation}
Note that before the first observation, $k < k_0$, $x_T(k) = x(k_0)$. After the last observation, $k \geq k_{|K|-1}$, $x_T(k) = x(k_{|K|-1})$. 
We make details of the point-preserving sampler and its usage in imputation in Algorithm \ref{alg:hybrid_sampler}.

\section{Related Works}

\paragraph{\textbf{Time Series Generation Models}}
Diffusion models outperform in time series synthesis tasks, compared to models based on VAE \cite{naiman2024generative}, GANs \cite{yoon2019time}. \citet{rasul2021autoregressive} initially used diffusion networks for time series based on recurrent networks. For imputation, SSSD \cite{lopez2023diffusionbased} and CSDI \cite{tashiro2021csdi} considered time series imputation similar to image inpainting tasks. TimeDiff \cite{shen2023non}, LDT \cite{feng2024latent}, and TMDM \cite{li2024transformermodulated} focused on time series forecasting tasks. \citet{coletta2024constrained} investigated guiding the sampling with a penalty function called DiffTime.
Our network design follows Diffusion-TS \cite{yuan2024diffusionts}, which introduced a decomposition architecture to disentangle time series data into trend and seasonal components and applied a Fourier-based loss term. Recent studies also explore frequency-domain diffusion \cite{crabbe2024time} and transformer-based alternatives \cite{chen2024sdformer}.


\paragraph{\textbf{Schrödinger Bridge in Time Series}}
Several studies have brought Schrödinger bridges to time series tasks. \citet{chen2023provably} analyzed algorithmic convergence and applied the bridge to imputation. \citet{hamdouche2023generative} combined kernel-based bridges with feed-forward and LSTM networks for deep hedging.
\citet{garg2024soft} relaxed the Schrödinger bridge by setting a geometric mixture for the prior distribution and applied their method to time series.
All of these methods require iterative solvers, which are computationally expensive and incompatible with standard diffusion pipelines. Also, none of them provides a unified framework for both unconditional and conditional generation. To the best of our knowledge, our paper is the first to use {diffusion bridge}, extending beyond classical Schrödinger bridge formulations and providing flexible priors.

\begin{algorithm}[!t]
    \caption{Point-preserving Sampler for TimeBridge}
    \label{alg:hybrid_sampler}
    \KwIn{Model \(D_\theta(\rvx_t)\), score function \(\rvs(D_\theta, \rvx_{t_i}, t_i, \rvx_{T}, {T})\), diffusion steps \(\{0 = t_0 < \cdots < t_\Gamma = T \}\),  Condition $\rvy$, masking \(\rvm\), step ratio \(s\)}
    \KwOut{Denoised output \(\rvx_0\)}
    $\rvx_{T} \leftarrow$ Set the prior as the hard constraint condition $\rvy$\\
    \lIf{Imputation }{$\rvx_{T} \leftarrow$ Linear spline interpolation  using \Eqref{eq:interpolate} with masking $\rvm$}
    \For{\(i = \Gamma, \ldots, 1\)}{
        \textbf{Sample} \(\boldsymbol{\rvvarepsilon} \sim \mathcal{N}(\mathbf{0, I})\)\\
        \(\tilde{t}_i \leftarrow t_i + s(t_{i-1} - t_i)\)\\
        {\footnotesize
        \(\rvd_{t_i} \leftarrow -\rvf(\rvx_{t_i}, t_i) + g^2(t_i)(\rvs(D_\theta, \rvx_{t_i}, t_i, \rvx_{T}, {T}) -  \rvh(\rvx_{t_i}, t_i, \rvx_T, T))\)}\\ 
        \(\tilde{\rvx}_{t_i} \leftarrow \rvx_{t_i} + \rvd_{t_i}\odot\rvm (\tilde{t}_i - t_{i}) + g(t_i)\sqrt{\tilde{t}_i - t_{i}} \cdot \boldsymbol{\rvvarepsilon}\odot\rvm\)\\
        \(\tilde{\rvd}_{t_i} \leftarrow -\rvf(\tilde{\rvx}_{t_i}, \tilde{t}_i) + g^2(\tilde{t}_i)(\frac{1}{2} \rvs(D_\theta, \tilde{\rvx}_{t_i}, \tilde{t}_i, \rvx_{T}, {T}) - \rvh(\tilde{\rvx}_{t_i}, \tilde{t}_i, \rvx_T, T))\)\\
        \(\rvx_{t_{i-1}} \leftarrow \tilde{\rvx}_{t_i} + \tilde{\rvd}_{t_i}\odot\rvm(t_{i-1} - \tilde{t}_i)\)\\
        \If{\(i \neq 1\)}{
            {\small\(\rvd_{t_{i}}' \leftarrow -\rvf(\rvx_{t_{i-1}}, t_{i-1}) + g^2(t_{i-1})  (\frac{1}{2} \rvs(D_\theta, \rvx_{t_{i-1}}, t_{i-1}, \rvx_{T}, {T}) - \rvh(\rvx_{t_{i-1}}, t_{i-1}, \rvx_T, T))\)}
            \\
            \(\rvx_{t_{i-1}} \leftarrow \tilde{\rvx}_{t_i} + \frac{1}{2}(\rvd_{t_i}' + \tilde{\rvd}_{t_{i}})\odot\rvm(t_{i-1} - \tilde{t}_i)\)
        }
    }
\end{algorithm}

\begin{table*}[!t]
\centering
\caption{Quality comparison of unconditional time series synthetic data with length of 24. The best results are in \textbf{bold} and the second best are \underline{underlined}. We report the average score (lower is better) and the rank of each method. The baseline experimental results (denoted by $\dagger$) are adopted from \cite{yuan2024diffusionts}. We re-implement Diffusion-TS with their official code.} 
\label{tab:main}
\resizebox{0.9\textwidth}{!}{%
\begin{tabular}{ccrrrrrrc}
\toprule
\multirow{2}{*}{Metric} & \multirow{2}{*}{Methods} & \multicolumn{2}{c}{Synthetic datasets} & \multicolumn{4}{c}{Real datasets} & \multirow{2}{*}{\begin{tabular}[c]{@{}c@{}}Average\\(rank)\end{tabular}} \\ \cmidrule(lr){3-4} \cmidrule(lr){5-8}
 &  & \multicolumn{1}{c}{Sines} & \multicolumn{1}{c}{MuJoCo} & \multicolumn{1}{c}{ETTh} & \multicolumn{1}{c}{Stocks} & \multicolumn{1}{c}{Energy} & \multicolumn{1}{c}{fMRI} & \\ \midrule
\multirow{8}{*}{\begin{tabular}[c]{@{}c@{}}Context-FID\\score\\(Lower is better)\end{tabular}} & TimeVAE$^\dagger$ & 0.307±.060 & 0.251±.015 & 0.805±.186 & 0.215±.035 & 1.631±.142 & 14.449±.969 & 2.943 (8) \\
 & TimeGAN$^\dagger$  & 0.101±.014 & 0.563±.052 & 0.300±.013 & 0.103±.013 & 0.767±.103 & 1.292±.218 & 0.521 (6) \\
 & Cot-GAN$^\dagger$ & 1.337±.068 & 1.094±.079 & 0.980±.071 & 0.408±.086 & 1.039±.028 & 7.813±.550 & 2.112 (7) \\
 & Diffwave$^\dagger$ & 0.014±.002 & 0.393±.041 & 0.873±.061 & 0.232±.032 & 1.031±.131 & 0.244±.018 & 0.465 (5) \\
 & DiffTime$^\dagger$ & \textbf{0.006±.001} & 0.188±.028 & 0.299±.044 & 0.236±.074 & 0.279±.045 & 0.340±.015 & 0.225 (4) \\
 & Diffusion-TS & \underline{0.008±.001} & \textbf{0.012±.002} & 0.129±.008 & 0.165±.043 & 0.095±.014 & 0.106±.007 & 0.086 (3) \\  \cmidrule(lr){2-8} \cmidrule(lr){9-9}
 & \textbf{TimeBridge} & \underline{0.008±.001} & 0.018±.002 & \underline{0.069±.004} & \underline{0.079±.023} & \underline{0.082±.007} & \underline{0.097±.008} & \underline{0.059 (2)} \\
 & \textbf{TimeBridge-GP} & \textbf{0.006±.001}  &  \underline{0.016±.002} & \textbf{0.067±.007} & \textbf{0.054±.009} & \textbf{0.064±.007} & \textbf{0.096±.008} & \textbf{0.051 (1)} \\  \hline
\multirow{8}{*}{\begin{tabular}[c]{@{}c@{}}Correlational\\score\\(Lower is better)\end{tabular}} & TimeVAE$^\dagger$ & 0.131±.010 & 0.388±.041 & 0.111±020 & 0.095±.008 & 1.688±.226 & 17.296±.526 & 3.285 (6) \\
 & TimeGAN$^\dagger$  & 0.045±.010 & 0.886±.039 & 0.210±.006 & 0.063±.005 & 4.010±.104 & 23.502±.039 & 4.786 (7) \\
 & Cot-GAN$^\dagger$ & 0.049±.010 & 1.042±.007 & 0.249±.009 & 0.087±.004 & 3.164±.061 & 26.824±.449 & 5.236 (8) \\
 & Diffwave$^\dagger$ & 0.022±.005 & 0.579±.018 & 0.175±.006 & 0.030±.020 & 5.001±.154 & 3.927±.049 & 1.622 (5) \\
 & DiffTime$^\dagger$ & \textbf{0.017±.004} & 0.218±.031 & 0.067±.005 & \textbf{0.006±.002} & 1.158±.095 & 1.501±.048 & 0.495 (4) \\
 & Diffusion-TS & \textbf{0.017±.006} & 0.193±.015 & 0.049±.012 & 0.011±.008 & \textbf{0.874±.163} & 1.151±.053 & 0.383 (3) \\ \cmidrule(lr){2-8} \cmidrule(lr){9-9}
 & \textbf{TimeBridge} & 0.024±.001 & \underline{0.179±.011} & \underline{0.035±.005} & 0.009±.009 & 1.064±.167 & \underline{0.914±.028} & \underline{0.371 (2)} \\
 & \textbf{TimeBridge-GP} & \underline{0.018±.003} & \textbf{0.173±.025} & \textbf{0.034±.004} & \underline{0.008±.008} & \underline{0.902±.218} & \textbf{0.839±.018} & \textbf{0.329 (1) }\\   \hline
\multirow{8}{*}{\begin{tabular}[c]{@{}c@{}}Discriminative\\score\\(Lower is better)\end{tabular}} & TimeVAE$^\dagger$ & 0.041±.044 & 0.230±.102 & 0.209±.058 & 0.145±.120 & 0.499±.000 & 0.476±.044 & 0.267 (7) \\
 & TimeGAN$^\dagger$  & 0.011±.008 & 0.238±.068 & 0.114±.055 & 0.102±.021 & 0.236±.012 & 0.484±.042 & 0.198 (5) \\
 & Cot-GAN$^\dagger$ & 0.254±.137 & 0.426±.022 & 0.325±.099 & 0.230±.016 & 0.498±.002 & 0.492±.018 & 0.371 (8) \\
 & Diffwave$^\dagger$  & 0.017±.008 & 0.203±.096 & 0.190±.008 & 0.232±.061 & 0.493±.004 & 0.402±.029 & 0.256 (6) \\
 & DiffTime$^\dagger$ & 0.013±.006 & 0.154±.045 & 0.100±.007 & 0.097±.016 & 0.445±.004 & 0.245±.051 & 0.176 (4) \\
  & Diffusion-TS & \underline{0.008±.012} & 0.009±.011 &0.074±.010 & 0.083±.044 & \textbf{0.126±.007} & 0.133±.024 & 0.072 (3) \\ \cmidrule(lr){2-8} \cmidrule(lr){9-9}
 & \textbf{TimeBridge} & 0.012±.003 & \underline{0.007±.008} & \textbf{0.052±.004} & \underline{0.052±.021} & 0.167±.003 & \textbf{0.077±.010} & \underline{0.061 (2)} \\
 & \textbf{TimeBridge-GP} & \textbf{0.002±.002} & \textbf{0.006±.003} & \textbf{0.052±.002} &\textbf{0.049±.014} & \underline{0.165±.009} & \underline{0.082±.033} &  \textbf{0.059 (1)} 
\\ \bottomrule
\end{tabular}%
}
\vspace{-0.15cm}
\end{table*}

\begin{table}[!ht]
\centering
\caption{Quality comparison of unconditional time-series synthetic data with a longer length of 64. Lower is better.}
\label{tab:long}
\vspace{-0.2cm}
\resizebox{0.49\textwidth}{!}{%
\begin{tabular}{l l r r r}
\toprule
\textbf{Dataset} & \textbf{Method} &
\textbf{C.-FID ($\downarrow$)} &
\textbf{Corr. ($\downarrow$)} &
\textbf{Disc. ($\downarrow$)} \\ \midrule
ETTh   & Diffusion-TS            & 0.235±.011 & 0.061±.010 & 0.078±.017 \\
       & \textbf{TimeBridge-GP}  & \textbf{0.110±.009} & \textbf{0.038±.011} & \textbf{0.045±.013} \\ \addlinespace[2pt]
Stocks & Diffusion-TS            & 0.948±.227 & 0.005±.005 & 0.149±.035 \\
       & \textbf{TimeBridge-GP}  & \textbf{0.693±.165} & \textbf{0.005±.003} & \textbf{0.090±.018} \\ \addlinespace[2pt]
Energy & Diffusion-TS            & \textbf{0.067±.010} & \textbf{0.442±.077} & \textbf{0.065±.011} \\
       & \textbf{TimeBridge-GP}  & 0.072±.010 & 0.548±.102 & 0.149±.005 \\ \addlinespace[2pt]
fMRI   & Diffusion-TS            & 3.662±.104 & 4.585±.068 & 0.245±.234 \\
       & \textbf{TimeBridge-GP}  & \textbf{0.527±.024} & \textbf{1.569±.028} & \textbf{0.211±.184} \\\hline
Average & Diffusion-TS              & 1.228 & 1.273 & 0.134 \\
& \textbf{TimeBridge-GP} & \textbf{0.351} & \textbf{0.540} & \textbf{0.124} \\

\bottomrule
\end{tabular}%
}
\vspace{-0.3cm}
\end{table}

\section{Experiments}
\label{sec:exp}

\subsection{Experimental Setup}
We assess the performance using widely used time series datasets. We use two simulation datasets: \textbf{Sines} with 5 features of different frequencies and phases of sine functions and  \textbf{MuJoCo} of a multivariate advanced physics simulation with 14 features. For real-world datasets, we use \textbf{ETT} (Electricity Transformer Temperature) for long-term electric power with 7 features, \textbf{Stocks} of Google stock prices and volumes with 6 features, \textbf{Energy}, UCI dataset with appliances energy use in a low energy building with 28 features, and \textbf{fMRI} of the blood oxygen level-dependent functional MR imaging with 50 selected features.

For measures, we assess three metrics for unconditional time series generation on a normalized (0-1) scale. Context-Fréchet Inception Distance (Context-FID, C.-FID) score \cite{jeha2022psa} measures the distributional quality for mean and variance projected on embedding on trained TS2Vec \cite{yue2022ts2vec}. Correlational (Corr.) score \cite{liao2020conditional} measures temporal dependency of real and synthetic data using the absolute error of cross-correlation. Discriminative  (Disc.) score \cite{yoon2019time} measures the performance of a classifier trained to distinguish original and synthetic data in a supervised manner. For imputation, we use the mean squared error (MSE) and mean absolute error (MAE).

Our experiments are conducted using PyTorch on four NVIDIA GeForce RTX 4090 GPUs. We report the mean and its interval of five runs. For the details, refer to \Appref{app:exp}.

\begin{table*}[!ht]
\centering
\caption{Quality comparison of trend-conditioned time series synthetic data.} 
\label{tab:main-trend}
\vspace{-0.2cm}
\resizebox{0.95\textwidth}{!}{%
\begin{tabular}{ccrrrrrrc}
\toprule
\multirow{2}{*}{Metric} & \multirow{2}{*}{Methods} & \multicolumn{3}{c}{ETTh} & \multicolumn{3}{c}{Energy} & \multirow{2}{*}{\begin{tabular}[c]{@{}c@{}}Average\\(rank)\end{tabular}} \\ \cmidrule(lr){3-5} \cmidrule(lr){6-8}
 &  & \multicolumn{1}{c}{Linear} & \multicolumn{1}{c}{Polynomial} & \multicolumn{1}{c}{Butterworth} & \multicolumn{1}{c}{Linear} & \multicolumn{1}{c}{Polynomial} & \multicolumn{1}{c}{Butterworth} & \\ \midrule
\multirow{4}{*}{\begin{tabular}[c]{@{}c@{}}Context-FID\\score\end{tabular}} &
Trend Baseline & 0.1729±.0228 & 0.1000±.0790 & 0.7343±.6795 & 0.8343±.0531 & 0.5864±.2016 & 0.8093±.2690  & 0.5395 (-) \\
 & SSSD &  0.0137±.0028 & 0.0281±.0136 & 0.0575±.0104 & 0.0146±.0085 & 0.0135±.0015 & 0.0075±.0057   & 0.0225 (3) \\
 & Diffusion-TS& 0.0049±.0008 & {0.0028±.0003} & 0.0030±.0003 & 0.0052±.0006 & 0.0050±.0005 & 0.0024±.0003 & 0.0039 (2) \\
 & \textbf{TimeBridge} & \textbf{0.0028±.0003} & \textbf{0.0024±.0002} & \textbf{0.0023±.0002} & \textbf{0.0031±.0002} & \textbf{0.0036±.0002 }& \textbf{0.0017±.0002} & \textbf{0.0027 (1)} \\ \hline
\multirow{4}{*}{\begin{tabular}[c]{@{}c@{}}Correlational\\score\end{tabular}} & 
Trend Baseline & 0.0493±.0104 & 0.0372±.0139 & 0.0524±.0194 & 0.4816±.0633 & 0.4816±.0999  & 0.5246±.0915  & 0.2711 (-) \\
& SSSD & 0.0362±.0137 & 0.0334±.0074 & 0.0404±.0032 & 0.6186±.1124 & 0.5594±.0925 & 0.5138±.0917  & 0.3003 (3) \\
 & Diffusion-TS & 0.0247±.0109 & \textbf{0.0238±.0113} & 0.0239±.0127 & 0.4834±.2090 & 0.4843±.2116 & \textbf{0.4816±.2093} & 0.2536 (2) \\
 & \textbf{TimeBridge} & \textbf{0.0240±.0120} & \textbf{0.0238±.0124} & \textbf{0.0237±.0123} & \textbf{0.4828±.2101} & \textbf{0.4830±.2106} & 0.4833±.2071 & \textbf{0.2534 (1)} \\ \hline
\multirow{4}{*}{\begin{tabular}[c]{@{}c@{}}Discriminative\\score\end{tabular}} & Trend Baseline & 0.1258±.0795  & 0.0574±.0626 & 0.1380±.0943 & 0.5000±.0633 & 0.4996±.0004 & 0.4997±.0003  & 0.3034 (-) \\
& SSSD & 0.0272±.0105 &  0.0301±.0142 &  0.1108±.0915  & 0.1871±.0749 & 0.2198±.0385 & 0.1360±.0797 & 0.1185 (3) \\
 & Diffusion-TS & {0.0045±.0038} & {0.0038±.0040} & {0.0043±.0045} & \textbf{0.0031±.0035} & {0.0067±.0044} & {0.0046±.0034} & 0.0045 (2) \\
 & \textbf{TimeBridge} & \textbf{0.0036±.0038}  & \textbf{0.0023±.0021}  & \textbf{0.0021±.0022} & {0.0054±.0026} & \textbf{0.0056±.0048}  & \textbf{0.0037±.0035} & \textbf{0.0038 (1)}
\\ \bottomrule
\end{tabular}
}
\vspace{-0.15cm}
\end{table*}

\begin{table*}[t]
\centering
\caption{Imputation results of MSE in the order of 1e-3 on the Mujoco dataset with a random mask of ratio \{0.7, 0.8, 0.9\}. The best results are in \textbf{bold} and the second best are \underline{underlined}. The baseline experimental results are adopted from \cite{lopez2023diffusionbased} and \cite{yuan2024diffusionts}.} 
\label{tab:main-mujoco}
\vspace{-0.15cm}
\resizebox{0.8\textwidth}{!}{%
\begin{tabular}{crrrrrrrrrr}
\toprule
Methods & \multicolumn{1}{c}{\begin{tabular}[c]{@{}c@{}}RNN \\ GRU-D\end{tabular}} & \multicolumn{1}{c}{\begin{tabular}[c]{@{}c@{}}ODE\\ -RNN\end{tabular}} & \multicolumn{1}{c}{\begin{tabular}[c]{@{}c@{}}Neural\\ CDE\end{tabular}} & \multicolumn{1}{c}{\begin{tabular}[c]{@{}c@{}}Latent\\ -ODE\end{tabular}} & \multicolumn{1}{c}{NAOMI} & \multicolumn{1}{c}{NRTSI} & \multicolumn{1}{c}{CSDI} & \multicolumn{1}{c}{SSSD} & \multicolumn{1}{c}{\begin{tabular}[c]{@{}c@{}}Diffusion\\ -TS\end{tabular}} & \multicolumn{1}{c}{\textbf{TimeBridge}} \\ \midrule
70\% & 11.34 & 9.86 & 8.35 & 3 & 1.46 & 0.63 & \underline{0.24±.03} & 0.59±.08 & 0.37±.03 & \textbf{0.19±.00}  \\
80\% & 14.21 & 12.09 & 10.71 & 2.95 & 2.32 & 1.22 & 0.61±.10 & 1.00±.05 & \underline{0.43±.03} & \textbf{0.26±.02}  \\
90\% & 19.68 & 16.47 & 13.52 & 3.6 & 4.42 & 4.06 & 4.84±.02 & 1.90±.03 & \underline{0.73±.12} & \textbf{0.60±.02}
\\ \bottomrule
\end{tabular}%
}
\vspace{-0.15cm}
\end{table*}

\begin{table*}[t]
\centering
\caption{Imputation result of MSE and MAE on the ETTh and Energy datasets with a geometric mask of ratio \{0.25, 0.5, 0.75\}. The best results are in \textbf{bold} and the second best are \underline{underlined}.}
\label{tab:main-imputation}
\vspace{-0.2cm}
\resizebox{0.94\textwidth}{!}{%
\begin{tabular}{ccrrrcrrrc}
\toprule
\multicolumn{2}{c}{Metric} & \multicolumn{4}{c}{MSE} & \multicolumn{4}{c}{MAE} \\ 
\cmidrule(lr){3-6} \cmidrule(lr){7-10}
\multicolumn{2}{c}{Methods} & \multicolumn{1}{c}{SSSD} & \multicolumn{1}{c}{Diffusion-TS} & \multicolumn{1}{c}{\textbf{TimeBridge}} & \multicolumn{1}{c}{\textbf{TimeBridge-75\%}} & \multicolumn{1}{c}{SSSD} & \multicolumn{1}{c}{Diffusion-TS} & \multicolumn{1}{c}{\textbf{TimeBridge}} & \multicolumn{1}{c}{\textbf{TimeBridge-75\%}} \\ 
\midrule
\multirow{3}{*}{ETTh} & 25\% & 0.496±0.049 & 0.406±0.006 & \underline{0.159±0.014} & \textbf{0.146±0.007} & 0.433±0.009 & 0.378±0.001 & \textbf{0.220±0.001} & \textbf{0.220±0.002} \\
 & 50\% & 0.523±0.061 & 0.561±0.010 & 0.226±0.010 & \textbf{0.199±0.006} & 0.435±0.013 & 0.422±0.001 & 0.248±0.002 & \textbf{0.237±0.002} \\
 & 75\% & \textbf{0.588±0.012} & 0.774±0.006 & \underline{0.661±0.031} & \underline{0.661±0.031} & 0.449±0.001 & 0.497±0.001 & \textbf{0.366±0.004} & \textbf{0.366±0.004} \\ 
\midrule
\multirow{3}{*}{Energy} & 25\% & 108.82±19.12 & 137.77±1.14 & \underline{17.06±2.34} & \textbf{12.03±0.40} & 1.94±0.19 & 1.39±0.00 & \underline{0.47±0.01} & \textbf{0.45±0.00} \\
 & 50\% & 107.68±29.14 & 163.86±1.02 & \underline{30.64±3.00} & \textbf{23.07±2.13} & 1.77±0.10 & 1.70±0.01 & \underline{0.71±0.02} & \textbf{0.63±0.01} \\
 & 75\% & 139.07±4.49 & 199.38±1.82 & \textbf{63.84±2.15} & \textbf{63.84±2.15} & 1.87±0.01 & 2.09±0.01 & \textbf{1.07±0.01} & \textbf{1.07±0.01} \\
\bottomrule
\end{tabular}%
}
\vspace{-0.15cm}
\end{table*}

\subsection{Unconditional Generation}

We demonstrate the unconditional generation tasks with a length of 24 as baselines. We compare with TimeVAE \cite{desai2021timevae}, TimeGAN \cite{yoon2019time}, Cot-GAN \cite{xu2020cot}, Diffwave \cite{kong2021diffwave}, DiffTime \cite{coletta2024constrained} with SSSD \cite{lopez2023diffusionbased} backbone, and Diffusion-TS \cite{yuan2024diffusionts}. For the proposed TimeBridge, we choose the data-dependent settings in \Eqref{eq:mean}
and TimeBridge-GP denotes the time-dependent GP prior in \Eqref{eq:gp_prior}. As the prior mean and variance are derived from the training datasets, the proposed method does not require additional information. 

Results are shown in \Tabref{tab:main}. We observe that the proposed method outperforms the previous methods. The performance gain is significant in Context-FID and discriminative scores, reducing the scores of 40.70\% and 18.05\% on average, respectively.
Notably, the strength of data and time-dependent prior is shown in real datasets, such as ETTh and Stocks.
This indicates that our prior selection and modeling help in approximating the data distribution during synthesis.
In average rank, the proposed method shows the best performance. It indicates the power of prior selection in time series generation within a similar structure.
The results in \Tabref{tab:long} show that TimeBridge-GP produces higher-quality samples than standard Gaussian for long-horizon generation tasks with a sequence length of 64.

\begin{figure}[!t]
    \centering
    \begin{subfigure}{0.13\textwidth}
        \includegraphics[width=\linewidth]{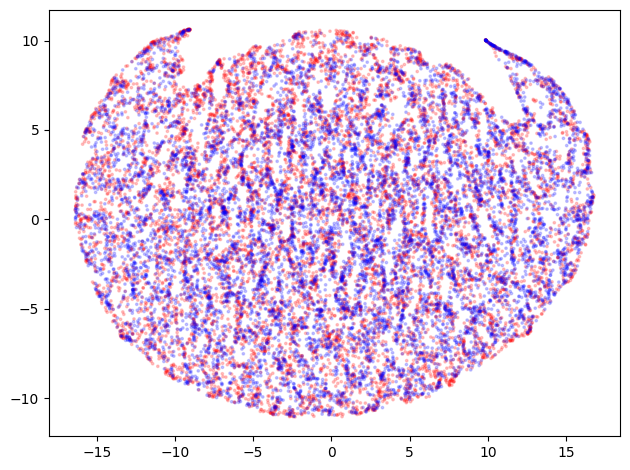}
        \caption{Sines}
        \label{fig:sub-sines}
    \end{subfigure}
    \begin{subfigure}{0.13\textwidth}
        \includegraphics[width=\linewidth]{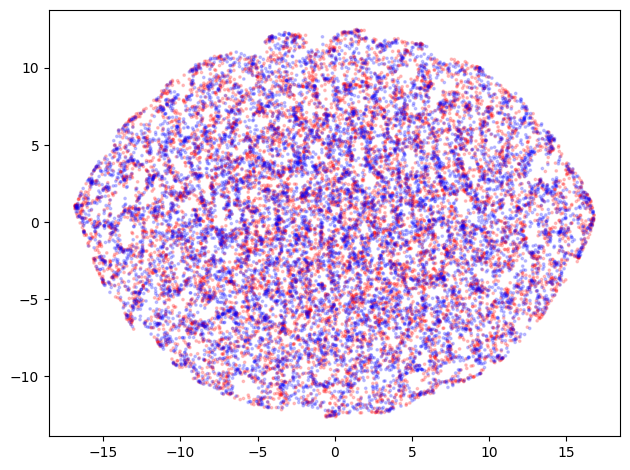}
        \caption{Mujoco}
        \label{fig:sub-mujoco}
    \end{subfigure}
    \begin{subfigure}{0.13\textwidth}
        \includegraphics[width=\linewidth]{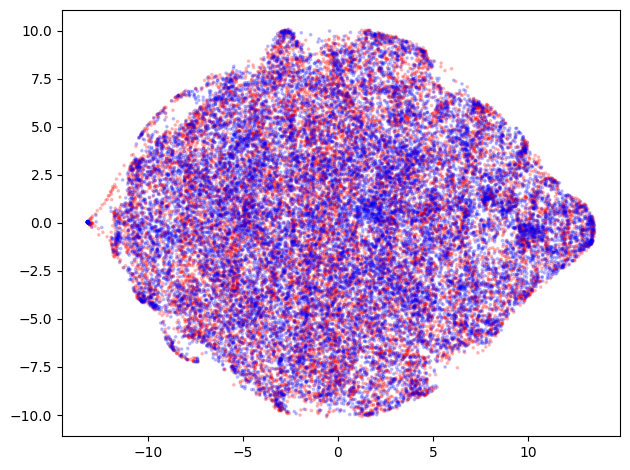}
        \caption{ETTh}
        \label{fig:sub-etth}
    \end{subfigure}
    \begin{subfigure}{0.13\textwidth}
        \includegraphics[width=\linewidth]{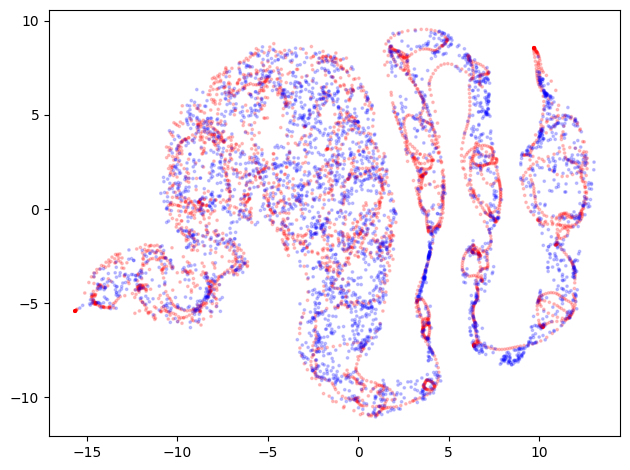}
        \caption{Stocks}
        \label{fig:sub-stocks}
    \end{subfigure}
    \begin{subfigure}{0.13\textwidth}
        \includegraphics[width=\linewidth]{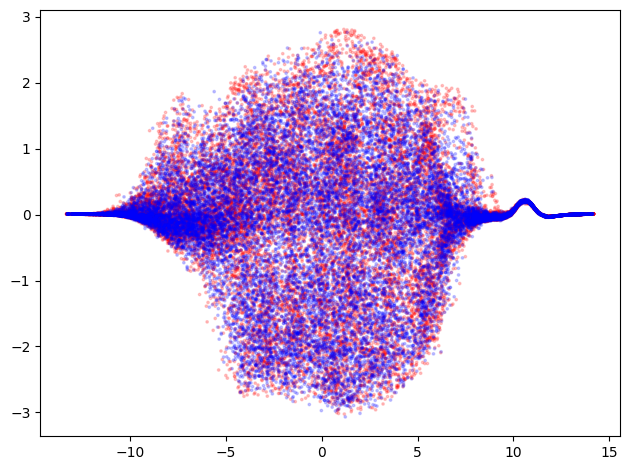}
        \caption{Energy}
        \label{fig:sub-energy}
    \end{subfigure}
    \begin{subfigure}{0.13\textwidth}
        \includegraphics[width=\linewidth]{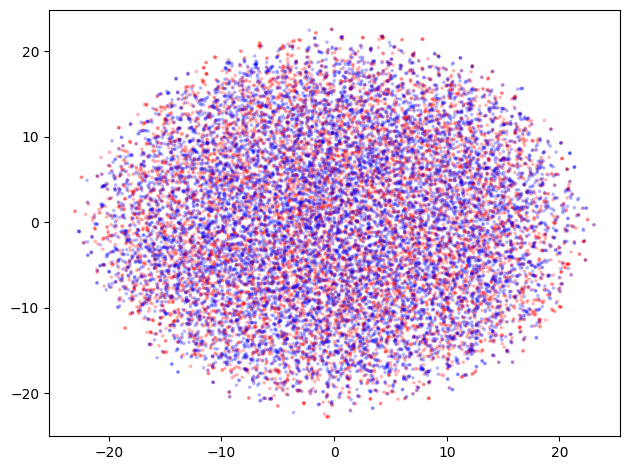}
        \caption{fMRI}
        \label{fig:sub-fmri}
    \end{subfigure}
    \caption{Visualization on t-SNE. Red and blue indicate original and synthetic data, respectively.}
    \label{fig:visualization-tsne}
    \vspace{-0.5cm}
\end{figure}

For visualization, we show the t-SNE plots with the generated data in \Figref{fig:visualization-tsne}. The red original data and the blue synthetic data show an overlapped plot in almost all datasets. The generated samples well-approximate the distribution properties in complex datasets, such as the Stocks and Energy datasets.


\subsection{Synthesis with Trend Condition}
We now consider trend-guided generation, where the data samples are in $\mathcal{D} = \{(\rvx^i, \rvy^i)\}_{i=1}^{N}$. For the trend $\rvy$, we consider three types: linear, polynomial with degree of three, and Butterworth filter \cite{butterworth1930theory}, a signal processing filter that captures trends \cite{xia2024market}. Using these trends as conditions, we reimplement the experiments in \Tabref{tab:main} with the ETTh and Energy datasets. To apply the conditions to baseline methods, we feed the trend into a conditional embedding and concatenate it with the original embedding. For TimeBridge, we set the prior $\rvx_T = \rvy$ upon the conditional embedding.

Experimental results are shown in \Tabref{tab:main-trend}. The rows with trend baseline indicate the results with the trends without any synthesis. After synthesis with trend conditions, we observe that the quality metrics diminished compared to the unconditional generation in \Tabref{tab:main}, indicating that trend conditions help the model generate more realistic data samples. With the trend condition in $\rvx_T$, the proposed TimeBridge achieves the best performance in Context-FID in all settings and the best discriminative score in 5 out of 6 settings while maintaining correlation scores at similar levels to Diffusion-TS. This indicates that the proposed framework can be broadly adopted by different conditioning settings, without controlling any details other than the prior distribution. The sampling path of TimeBridge from trend to data is shown in \Figref{fig:path_a}.

\subsection{Imputation with Fixed Masks}
For imputation, we compare the proposed method with imputation diffusion models and the conditional Langevin sampler of Diffusion-TS \cite{yuan2024diffusionts}. We use random masks for the Mujoco dataset and the geometric mask from \cite{zerveas2021transformer} for the ETTh and Energy datasets. 
We set the total length of imputation to 100 for the Mujoco dataset, matching the data length, and split the ETTh and Energy datasets into 48 time steps. For the proposed TimeBridge, we preserve the observed values and set the prior $\rvx_T$ as the interpolated values of the condition $\rvc$ as in \Eqref{eq:interpolate}.

\begin{figure*}[!t]
    \centering
    \begin{subfigure}{0.8\textwidth}
        \centering
        \includegraphics[width=0.98\linewidth]{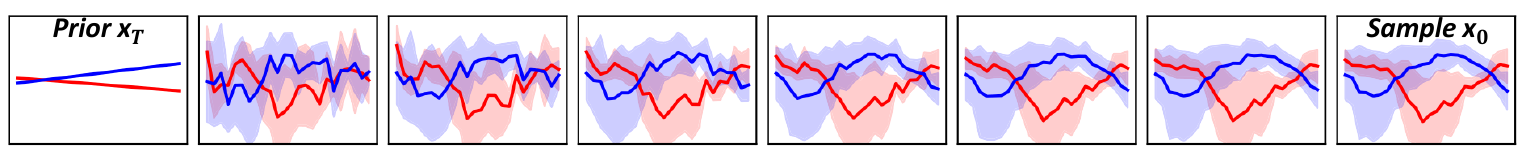}
        \includegraphics[width=\linewidth]{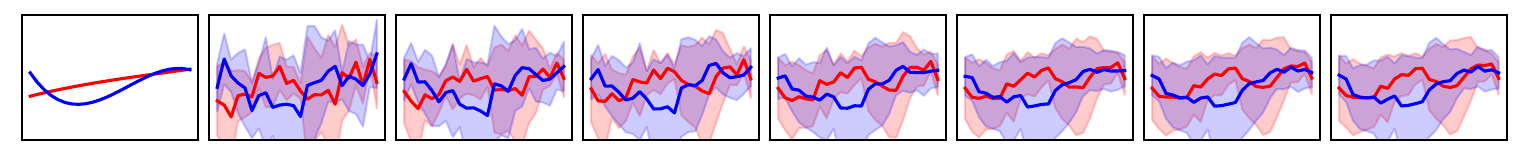}
        \vspace{-0.56cm}
        \caption{Sampling of TimeBridge by applying the two pair-wise trends (red/blue) as priors.}
        \label{fig:path_a}
    \end{subfigure}
    \begin{subfigure}{0.8\textwidth}
        \centering
        \includegraphics[width=0.98\linewidth]{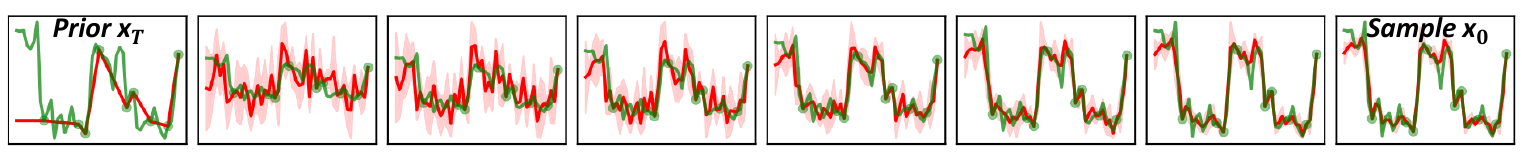}
        \includegraphics[width=\linewidth]{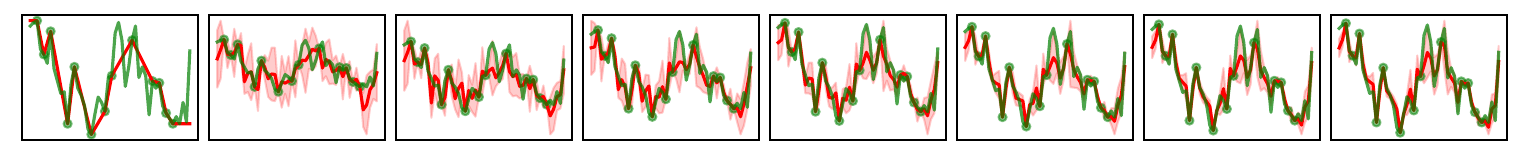}
        \vspace{-0.6cm}
        \caption{Imputation (red) of TimeBridge with point-preserving sampling using observed values (dots) of truth (green).}
        \label{fig:path_b}
    \end{subfigure}
    \vspace{-0.3cm}
    \caption{Illustration of sampling path from (Left) prior to (Right) data samples on the ETTh.}
    \label{fig:path}
    \vspace{-0.3cm}
\end{figure*}

\begin{figure}[!t]
    \centering
    \begin{subfigure}{0.4\textwidth}
        \centering
        \includegraphics[width=\linewidth]{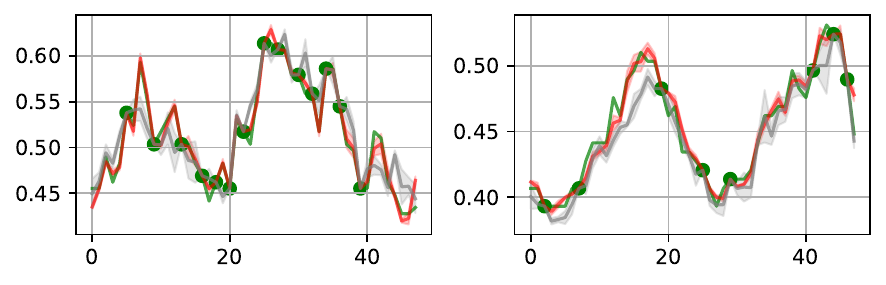}
        \vspace{-0.7cm}
        \caption{ETTh Imputation}
        \label{fig:etth_imputation}
    \end{subfigure}
    \begin{subfigure}{0.4\textwidth}
        \centering
        \includegraphics[width=\linewidth]{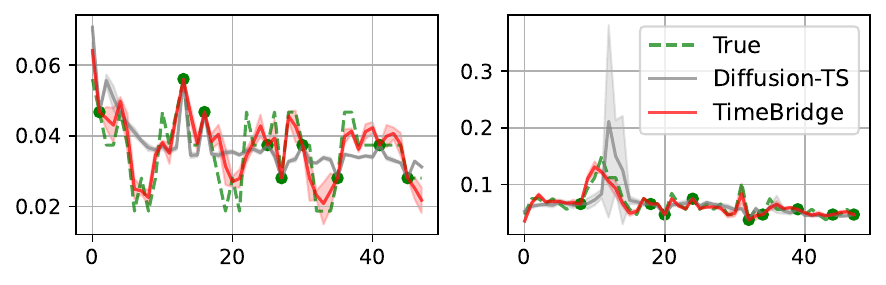}
        \vspace{-0.7cm}
        \caption{Energy Imputation}
        \label{fig:energy_imputation}
    \end{subfigure}
    \vspace{-0.2cm}
    \caption{Visualization of imputation on (a) ETTh and (b) Energy dataset. The green dotted lines and circles denote true and observed values with their intervals, respectively.}
    \label{fig:visualization}
\vspace{-0.4cm}
\end{figure}

The imputation results are shown in Tables \ref{tab:main-mujoco} and \ref{tab:main-imputation}. Both tables demonstrate that TimeBridge achieves the lowest measures in imputation tasks. Since the Langevin sampler of Diffusion-TS does not require individual training for each missing ratio, we ensure a fair comparison by evaluating TimeBridge with the same training setting, TimeBridge-75\%, which uses 75\% missing values for training. TimeBridge-75\% even outperforms the models trained with equal missing values, highlighting the generalization capability of TimeBridge in imputation tasks. The visualization of imputation results and sampling path can be found in Figures \ref{fig:visualization} and \ref{fig:path_b}, respectively.

\begin{table}[!t]
\center
\caption{Ablation study for training and sampling features.} 
\label{tab:ablation}
\vspace{-0.2cm}
\resizebox{0.45\textwidth}{!}{%
\begin{tabular}{ccrr}
\toprule
\multirow{2}{*}{Metric} & \multirow{2}{*}{Methods} & \multicolumn{2}{c}{Datasets} \\ \cline{3-4} 
 &  & \multicolumn{1}{c}{ETTh} & \multicolumn{1}{c}{Energy} \\ \midrule
\multirow{4}{*}{\begin{tabular}[c]{@{}c@{}}Context-FID \\ score\\(Lower is better)\end{tabular}}
 & TimeBridge-GP & \textbf{0.067±.007} &  \textbf{0.064±.007}  \\ \cline{2-4}
 & VE scheduler & 0.256±.028 & 0.184±.003  \\
 & F-matching & 0.087±.010 & 0.081±.006 \\ 
 & w/o Fourier & 0.087±.010 & 0.093±.011 \\\hline
\multirow{4}{*}{\begin{tabular}[c]{@{}c@{}}Discriminative score\\(Lower is better)\end{tabular}} 
 & TimeBridge-GP & \textbf{0.052±.002} & \textbf{0.165±.009} \\ \cline{2-4}
 & VE scheduler & 0.089±.009 & 0.236±.012 \\
 & F-matching & \textbf{0.052±.012} & 0.170±.007 \\
 & w/o Fourier & 0.060±.004 & 0.216±.007 \\
\bottomrule
\end{tabular}%
}
\vspace{-0.3cm}
\end{table}

\subsection{Additional Experiments}
\paragraph{\textbf{Ablation Study}}
\Tabref{tab:ablation} evaluates the effectiveness of training design with three variants: (1) variance exploding (VE) scheduler, (2) noise matching instead of directly predicting $D_\theta$ as in \citet{zhou2024denoising}, and (3) without Fourier-based loss during training. The results indicate that each component of our diffusion scheduler and loss design contributes to improved performance.

\paragraph{\textbf{Wasserstein distance}}

To experimentally justify the strength of the diffusion prior, we measure the Wasserstein distance within the diffusion bridge’s optimal transport framework. \citet{kollovieh2024flow} argue that closer alignment of data and prior distributions reduces path complexity, and time-reflective priors can lower the Wasserstein distance. As shown in \Tabref{tab:wasserstein}, our data-dependent priors (unconditional and imputation) yield shorter distances. While not always a direct indicator of performance, this may explain why diffusion bridge models surpass standard diffusion approaches.

\begin{table}[!t]
\centering
\caption{Wasserstein distance results for unconditional and imputation settings using different priors.}
\label{tab:wasserstein}
\vspace{-0.2cm}
\resizebox{0.45\textwidth}{!}{%
\begin{tabular}{@{}lcccc@{}}
\toprule
 & \multicolumn{2}{c}{{Unconditional}} & \multicolumn{2}{c}{{Imputation}} \\ 
\cmidrule(lr){2-3} \cmidrule(lr){4-5}
{Prior} & $\mathcal{N}(\mathbf{0}, \mathbf{I})$ & $\mathcal{N}(\boldsymbol{\mu}, \boldsymbol{\Sigma})$ (Ours) & $\mathcal{N}(\mathbf{0}, \mathbf{I})$ & Linear Spline (Ours) \\ 
\midrule
ETTh & 14.285 & \textbf{2.387} & 8.000 & \textbf{0.485} \\ 
Energy & 28.128 & \textbf{6.376} & 16.113 & \textbf{1.494} \\ 
\bottomrule
\end{tabular}
}
\vspace{-0.3cm}
\end{table}

\paragraph{\textbf{Extreme cases}}
To evaluate effectiveness on extreme values, we perform an ablation study on tailed data samples using real-world data in our experiments, i.e., ETTh, Stocks, and Energy datasets. For each real and synthetic dataset, we extract top $k\%$ data samples with high variance as extreme data, respectively. The results from samples with the top $20\%$ and $50\%$ variance are shown in \Tabref{tab:extreme}. Our method outperformed 10 out of 12 metrics, showing its robustness in high variance and tailed data.

\begin{table}[!t]
\centering
\caption{Ablation results on extreme values of top $k\%$ variance.}
\label{tab:extreme}
\vspace{-0.2cm}
\resizebox{0.49\textwidth}{!}{%
\begin{tabular}{@{}llcccccc@{}}
\toprule
{Top $k\%$} & {Metric} & \multicolumn{2}{c}{{ETTh}} & \multicolumn{2}{c}{{Energy}} & \multicolumn{2}{c}{{Stock}} \\ 
\cmidrule(lr){3-4} \cmidrule(lr){5-6} \cmidrule(lr){7-8}
 &  & Diffusion-TS & Ours & Diffusion-TS & Ours & Diffusion-TS & Ours \\ 
\midrule
20\% & C.-FID & 0.153 & \textbf{0.103} & 0.128 & \textbf{0.093} & 0.788 & \textbf{0.257} \\ 
     & Corr. & 0.095 & \textbf{0.070} & 0.946 & \textbf{0.150} & 1.217 & \textbf{0.010} \\ 
50\% & C.-FID & 0.144 & \textbf{0.098} & 0.139 & \textbf{0.096} & 0.475 & \textbf{0.154} \\ 
     & Corr. & 0.062 & \textbf{0.048} & 0.886 & \textbf{0.021} & 1.028 & \textbf{0.012} \\ 
\bottomrule
\end{tabular}
}
\vspace{-0.3cm}
\end{table}

\paragraph{\textbf{Longer Sequences}}
Modeling longer sequences is crucial for time series generation. Thus, we have tested lengths of 192 and 360, common for forecasting cyclic or annual patterns. By controlling only $\sigma_{data}$ and fixing other parameters to minimize the search space, \Tabref{tab:seq_len_comparison} shows that our method outperforms standard diffusion priors, excelling in 11 out of 12 metrics. 
\begin{table}[!t]
\centering
\caption{Comparison of longer sequences of 192 and 360.}
\label{tab:seq_len_comparison}
\vspace{-0.2cm}
\resizebox{0.49\textwidth}{!}{%
\begin{tabular}{@{}llrrrrrr@{}}
\toprule
 &  & \multicolumn{2}{c}{Context\textnormal{-}FID} & \multicolumn{2}{c}{Cross\textnormal{-}Correlation} & \multicolumn{2}{c}{Discriminative} \\
\cmidrule(lr){3-4} \cmidrule(lr){5-6} \cmidrule(lr){7-8}
Dataset (length) &  & Diffusion-TS & Ours & Diffusion-TS & Ours & Diffusion-TS & Ours \\
\midrule
Stocks (192) &  & 0.679 & \textbf{0.548} & 0.009 & \textbf{0.006} & 0.144 & \textbf{0.060} \\
Stocks (360) &  & 0.954 & \textbf{0.606} & 0.016 & \textbf{0.004} & 0.125 & \textbf{0.069} \\[2pt]
ETTh\;(192)  &  & \textbf{1.021} & 1.535 & 0.106 & \textbf{0.098} & 0.187 & \textbf{0.096} \\
ETTh\;(360)  &  & 1.977 & \textbf{1.780} & 0.114 & \textbf{0.099} & 0.226 & \textbf{0.109} \\
\bottomrule
\end{tabular}%
}
\vspace{-0.3cm}
\end{table}
\noindent


\paragraph{\textbf{Quadratic Imputation Prior}}
Instead of the linear interpolation used in \Tabref{tab:main-imputation}, we compared using quadratic interpolation methods in \Tabref{tab:interp}. Quadratic interpolation reduces MSE in several missing-rate scenarios (25\%, 50\%, 75\%), while linear interpolation remains competitive and more stable overall. We exclude the smoothing-based variant because it incurs 10$\times$ higher computation and undesirably smooths fixed observed points.

\subsection{Computation efficiency}
\Tabref{tab:compute_timebridge} illustrates the computational time for constructing priors and training. For per-iteration overhead, one iteration of our method requires only $1.06\times$ the computation of Diffusion-TS. Moreover, during sampling, our sampler uses fewer NFEs and runs significantly faster.

Prior construction involves extracting only the empirical mean and variance from 18K training samples, which is comparable to model initialization time. Imputation priors require slightly more computation but remain marginal relative to full training cost. Linear spline priors are further accelerated using the Numba\footnote{\url{https://github.com/numba/numba}} package with JIT compilation.

\begin{table}[!t]
\centering
\caption{Comparison of linear and quadratic interpolation priors (MSE at different missing ratios).}
\label{tab:interp}
\vspace{-0.2cm}
\resizebox{0.4\textwidth}{!}{
\begin{tabular}{lccc|ccc}
\toprule
\multirow{2}{*}{Dataset} & \multicolumn{3}{c}{Linear} & \multicolumn{3}{c}{Quadratic} \\
\cmidrule(lr){2-4} \cmidrule(lr){5-7}
& 25\% & 50\% & 75\% & 25\% & 50\% & 75\% \\
\midrule
ETTh   & 0.66  & 0.20  & 0.15  & 0.63  & 0.20  & 0.12 \\
Energy & 12.97 & 23.07 & 63.84 & 15.62 & 30.05 & 62.42 \\
\bottomrule
\end{tabular}
}
\vspace{-0.3cm}
\end{table}

\begin{table}[!t]
\centering
\caption{Detailed computation comparison for prior construction, training, and sampling.}
\label{tab:compute_timebridge}
\vspace{-0.2cm}
\resizebox{0.49\textwidth}{!}{
\begin{tabular}{lcccccc}
\toprule
& \multicolumn{3}{c}{\textbf{Prior Construction}} 
& \multicolumn{2}{c}{\textbf{Training}} 
& \textbf{Sampling} \\
\cmidrule(lr){2-4} \cmidrule(lr){5-6} \cmidrule(lr){7-7}
Method & Init & Uncond. & Imput. & Iter/step & Total & Time \\
\midrule
Diffusion-TS & 0.18s & -- & -- & 0.081s & 1458s & 90.14s \\
Ours & 0.18s & 0.09s & 2.90s & 0.086s (1.06$\times$) & 1548s (1.06$\times$) & 25.05s (0.28$\times$) \\
\bottomrule
\end{tabular}
}
\vspace{-0.3cm}
\end{table}


\section{Discussion and Conclusion}
\label{sec:conc}
In this paper, we highlight the effectiveness of selecting prior distributions for time series generation using diffusion bridges. To address the limitations of the standard Gaussian prior in diffusion models, we propose a framework that allows the selection of any prior suitable for various tasks.

While the results are encouraging, several limitations remain. First, we rely on the standard architecture inherited from diffusion time-series synthesizers, which might not fully exploit the benefits of diffusion bridges. Future work should investigate transformer-based designs that incorporate prior information more effectively.
Second, we leave the real-world applications with synthetic data beyond data synthesis itself for future study; realistic time-series synthesis could provide significant practical value.

As the proposed method can cover a wide range of scenarios in time series diffusion models, we believe this work significantly advances the investigation of prior distributions for developing a general time series generation framework. Ongoing advances in diffusion architectures will allow our approach to function as a plug-and-play adaptation that integrates with upcoming models.

\begin{acks}
This research was supported by the National Research Foundation of Korea (NRF) (No. RS-2024-00338859), and the Institute of Information \& communications Technology Planning \& Evaluation (IITP) (No. RS-2022-II220984) grants funded by the Korea government (MSIT). 
Jinseong Park is supported by a KIAS Individual Grant (AP102301) via the Center for AI and Natural Sciences at Korea Institute for Advanced Study. This work was supported by the Center for Advanced Computation at Korea Institute for Advanced Study.
\end{acks}

\clearpage


\bibliographystyle{ACM-Reference-Format}
\bibliography{bib}


\appendix

\section{Diffusion Bridge}
\label{app:diffusion_bridge}


\subsection{Schrödinger Bridge}
Standard diffusion models suffer from limitations such as the requirement of a sufficiently long time for the prior distribution to approximate a standard Gaussian distribution. To address these challenges, diffusion Schrödinger bridge models have been introduced \citep{de2021diffusion, chen2021likelihood} to find an entropic optimal transport between two probability distributions in terms of Kullback–Leibler divergence on path spaces \citep{chen2021likelihood, leonard2012schrodinger}. While standard diffusion models and flow matching models are not guaranteed to provide optimal transport, Schrödinger bridge models aim to find paths that recover entropy-regularized versions of optimal transport \citep{shi2024diffusion} as follows:
\begin{equation}
    \small
    \underset{p \in \mathcal{P}_{[0,T]}}{min}D_{KL}(p \lVert p^{ref}), \quad s.t. \ p_0 = p_{data}, \ p_T = p_{prior},
\end{equation}
where $\mathcal{P}_{[0,T]}$ refers to the space of path measures on $[0,T]$, $p^{ref}$ denotes the reference path measure, and $p_0,p_T$ represents the marginal distributions of $p$ at each time step $0,T$, respectively.
Setting the reference path as the forward SDE, the problem becomes equivalent to following forward-backward SDEs:

\begin{equation}
    \begin{split}
     d\rvx_t = [\rvf(\rvx_t,t)+g^2(t) \nabla \log \boldsymbol{\Psi}_t(\rvx_t)]dt + g(t) d\rvw_t,
    \end{split}
\end{equation}
\begin{equation}
    \begin{split}
    d\rvx_t = [\rvf(\rvx_t,t)-g^2(t) \nabla \log \hat{\boldsymbol{\Psi}}_t(\rvx_t)]dt + g(t) d\bar{\rvw}_t,
    \end{split}
\end{equation}

where $\nabla \log \boldsymbol{\Psi}_t (x_t) $ and $ \nabla \log \hat{\boldsymbol{\Psi}}_t (\rvx_t) $ are described as:

\begin{equation}
    \begin{split}
    & \frac{\partial\boldsymbol{\Psi}}{\partial t} = - \nabla_x \boldsymbol{\Psi}^T \rvf - \frac{1}{2} tr(g^2\nabla^2_x\boldsymbol{\Psi}) \\
    & \frac{\partial\hat{\boldsymbol{\Psi}}}{\partial t} = - \nabla_x \cdot (\hat{\boldsymbol{\Psi}}\rvf) + \frac{1}{2} tr(g^2\nabla^2_x\hat{\boldsymbol{\Psi}}) \\
    s.t. \quad \boldsymbol{\Psi}_0 \hat{\boldsymbol{\Psi}}_0 = &p_{data}, \boldsymbol{\Psi}_T  \hat{\boldsymbol{\Psi}}_T = p_{prior}, p_t = \boldsymbol{\Psi}_t \hat{\boldsymbol{\Psi}}_t \ \text{for} \ t \in [0,T].
    \end{split}
\end{equation}

The Schrödinger bridge problem can be addressed using iterative methods \citep{kullback1968probability}. At the outset, several early studies employed iterative methods to simulate trajectories converging to the Schrödinger bridge problem  \citep{vargas2021solving, shi2024diffusion, de2021diffusion}.
A classical solver for Schrödinger bridges is Iterative Proportional Fitting (IPF).
Beginning with arbitrary half-potentials $(\boldsymbol{\Psi}^{(0)}_t,\hat{\boldsymbol{\Psi}}^{(0)}_t)$, the algorithm alternates two normalization steps:
\begin{equation}
    \begin{split}
   \text{(Forward-update)}\quad
\hat{\boldsymbol{\Psi}}^{(n+1)}_t
  = \frac{p_{data}}{\int \boldsymbol{\Psi}^{(n)}_t\,d\rvx_0}, 
\\
   \text{(Backward-update)}\quad
\boldsymbol{\Psi}^{(n+1)}_t
  = \frac{p_{prior}}{\int \hat{\boldsymbol{\Psi}}^{(n+1)}_t\,d\rvx_T}. 
    \end{split}
\end{equation}

After each full iteration, the product $\boldsymbol{\Psi}^{(n+1)}_t\hat{\boldsymbol{\Psi}}^{(n+1)}_t$ matches both marginals, the sequence converges geometrically, requiring one forward–backward integration per iteration.
However, these methods rely on expensive iterative approximation techniques and have seen limited empirical application \citep{zhou2024denoising}.

\subsection{Denoising Diffusion Bridge Models}
Alternatively, \citet{chen2021likelihood} leveraged the theory of forward-backward SDEs to derive an exact log-likelihood expression for the Schrödinger bridge, which accurately generalizes the approach for score generative models. Drawing inspiration from the perspective that the framework of diffusion bridges, employing forward-backward SDEs, encompasses the paradigms of score matching diffusion models \citep{song2020denoising} and flow matching optimal transport paths \citep{lipman2022flow}. \citet{zhou2024denoising} demonstrated that reintroducing several design choices from these domains becomes feasible. In particular, the reparameterization outlined in \citet{karras2022elucidating} are utilized. 

\begin{table}[!t]
\centering
\vspace{-0.3cm}
\caption{VP and VE of diffusion bridge settings in DDBM.}
\label{tab:noise_sch}
\vspace{-0.2cm}
\resizebox{0.49\textwidth}{!}{%
    \begin{tabular}{cccccc}
    \toprule
          & $\rvf(\rvx_t,t)$ & $g^2(t)$ & $p(\rvx_t|\rvx_0)$ & $SNR_t$ & $\nabla_{\rvx_t}\text{log}p(\rvx_T|\rvx_t)$ \\
         \midrule
         VP & $\frac{d\text{log}\alpha_t}{dt} \rvx_t$ & $\frac{d}{dt}\sigma^2_t - 2\frac{d \text{log}\alpha_t}{dt}\sigma^2_t$ & $\mathcal{N}(\alpha_t \rvx_0, \sigma^2_t \mathbf{I})$ & $\alpha^2_t / \sigma^2_t$ & $\frac{(\alpha_t / \alpha_T)\rvx_T - \rvx_t}{\sigma^2_t (SNR_t / SNR_T - 1)}$ \\
         VE & 0 & $\frac{d}{dt}\sigma^2_t$ & $\mathcal{N}(\rvx_0, \sigma^2_t \mathbf{I})$ & $1/\sigma^2_t$ & $\frac{\rvx_T - \rvx_t}{\sigma^2_T - \sigma^2_t}$ \\
         \bottomrule
    \end{tabular}
}
\vspace{-0.2cm}
\end{table}

For noise scheduling, common options include variance-preserving (VP) and variance-exploding (VE) diffusion \citep{song2020denoising}.  We express the reparameterization for the VP and VE bridge of \citet{zhou2024denoising} in \Tabref{tab:noise_sch}. In our study, we chose to utilize the VP bridge, as our experimental findings indicate its superiority of VP over VE. For implementing the diffusion sampler, we modify the official code of DDBM \cite{zhou2024denoising} in \url{https://github.com/alexzhou907/DDBM}.

\section{Experimental Settings} 
\label{app:exp}

\begin{table*}[!t]
\centering
\caption{Hyperparameters for unconditional generation.}
\label{tab:uncond-setting}
\resizebox{0.9\textwidth}{!}{%
\begin{tabular}{lllcccccc}
\toprule
Setup & Hyper-parameter & Search space & \multicolumn{1}{c}{Sines} & \multicolumn{1}{c}{MuJoCo} & \multicolumn{1}{c}{ETTh} & \multicolumn{1}{c}{Stocks} & \multicolumn{1}{c}{Energy} & \multicolumn{1}{c}{fMRI} \\ \midrule
\multirow{5}{*}{Base setup \cite{yuan2024diffusionts}} 
 & Batch size & - & 128 & 128 & 128 & 64 & 64  & 64 \\
 & Training steps & - & 12000  & 14000 & 18000 & 10000  & 25000 & 15000 \\
 & Attention heads / head dim & - &  4/16  & 4/16 & 4/16 & 4/16  & 4/24 & 4/24\\
 & Layers of encoder / decoder & -  & 1/2  & 3/2 & 3/2 & 2/2  & 4/3 & 4/4\\
  & Warmup Learning rate & - & 0.008 & 0.008 & 0.008 & 0.008  & 0.008 & 0.008\\ 
 & Optimizer& - & Adam & Adam & Adam& Adam& Adam& Adam\\ 
 \midrule
 \multirow{3}{*}{TimeBridge} & $\beta_{min}$ & \{0.1, 0.2\} & 0.2 & 0.1 & 0.2 & 0.2 & 0.2 & 0.1 \\
  & $\beta_{d}$ & \{1, 2, 5, 10\} & 10 & 5 & 10 & 10 & 10 & 2 \\
  & $\sigma_{data}$ & \{0.05, 0.1, 0.5\} & 0.5 & 0.5 & 0.1 & 0.1 & 0.05 & 0.5 \\ 
  \midrule
  TimeBridge-GP & Kernel parameter $\eta$ & \{0.1, 0.3, 0.5, 1\} & 1 & 0.1 & 0.5 & 0.3 & 1 & 0.1 
\\ \bottomrule
\end{tabular}%
}
\end{table*}

\subsection{Baseline methods}
For unconditional generation, we compare with TimeVAE \cite{desai2021timevae}, TimeGAN \cite{yoon2019time}, Cot-GAN \cite{xu2020cot}, Diffwave \cite{kong2021diffwave}, DiffTime \cite{coletta2024constrained}, and Diffusion-TS \cite{yuan2024diffusionts}.
For \Tabref{tab:main}, we note the quality evaluation results in Diffusion-TS. We reimplement Diffusion-TS using the official GitHub in \url{https://github.com/Y-debug-sys/Diffusion-TS}, unfortunately, we cannot achieve the performance in the original paper. 
For conditional generation, we reimplement SSSD \cite{lopez2023diffusionbased} using the official GitHub \url{https://github.com/SSSD/sssd}.
For the trend-guided synthesis, we treat the trend as conditional observations and train the SSSD model to match the total time series data.
We add the conditional embedding layers to their original model.

\subsection{Evaluation Metrics}
FID score is a widely used metric in image tasks, capturing the similarity between generated images and real ones \cite{heusel2017gans}. To adapt it to the time series domain, \citet{jeha2022psa} proposed Context-FID score, which replaces the Inception model used in the original FID with TS2Vec, a pre-trained time series representation learning model \citep{yue2022ts2vec}. As Context-FID score aligns with the predictive accuracy of the generative model in downstream tasks, a lower Context-FID score indicates higher synthetic data quality. Both synthetic and real time-series data are encoded using a pre-trained TS2Vec model, and the FID score is calculated on the representations.

Correlational score is a metric used to evaluate the similarity in the covariance of time series features between original data and synthetic data \citep{liao2020conditional}. First, we estimate the covariance between the $i$-th and $j$-th features of the time series by:

\begin{equation}
    cov_{i,j} = \frac{1}{T} \sum_{t=1}^T X_i^t X_j^t - \Big{(} \frac{1}{T} \sum_{t=1}^T X_i^t \Big{)} \Big{(} \frac{1}{T} \sum_{t=1}^T X_j^t \Big{)}.
\end{equation}

\noindent
Then compute the correlation between two datasets as follows:

\begin{equation}
   \frac{1}{10} \sum_{i,j}^d \Big{\vert} \frac{cov_{i,j}^o}{\sqrt{cov_{i,i}^o cov_{j,j}^o}} - \frac{cov_{i,j}^s}{\sqrt{cov_{i,i}^s cov_{j,j}^s}} \Big{\vert},
\end{equation}

where $cov^o$ and $cov^s$ represent the covariance in the original data and synthetic data, respectively. 

Discriminative score is a metric designed to evaluate the performance of a classifier in distinguishing between real and synthetic data. The score is calculated as the absolute difference between the classifier's accuracy and 0.5.

\subsection{Additional Experimental Details}
\label{app:add_exp}

\paragraph{\textbf{Datasets}}

\begin{table}[!t]
\centering
\caption{Details of datasets.}
\label{tab:data-setting}
\resizebox{0.49\textwidth}{!}{%
\begin{tabular}{lccc}
\toprule
Dataset & Sample size & \# of features & Datatset link \\ \midrule
Sines & 10000 & 5 & https://github.com/jsyoon0823/TimeGAN \\ \hline
Stocks & 3773 & 6 & https://finance.yahoo.com/quote/GOOG \\ \hline
ETTh & 17420 & 7 & https://github.com/zhouhaoyi/ETDataset \\ \hline
MuJoCo & 10000 & 14 & https://github.com/google-deepmind/dm\_control \\ \hline
Energy & 19711 & 28 & https://archive.ics.uci.edu/dataset \\ \hline
fMRI & 10000 & 50 & https://www.fmrib.ox.ac.uk/datasets \\
\bottomrule
\end{tabular}%
}
\end{table}

\Tabref{tab:data-setting} presents the statistics of the datasets along with their online links.

\begin{table}[!t]
\centering
\caption{Hyperparameters for conditional generation. $\dagger$ We use 0.025 for the Energy with the Butterworth trend.}
\label{tab:cond-setting}
\resizebox{0.45\textwidth}{!}{%
\begin{tabular}{lllccc}
\toprule
\multirow{2}{*}{Method} & \multirow{2}{*}{Hyper-parameter}  & \multirow{2}{*}{Search space} & \multicolumn{3}{c}{Trend \& Imputation} \\ \cline{4-6}
 &  &   & \multicolumn{1}{c}{Mujoco} & \multicolumn{1}{c}{ETTh} & \multicolumn{1}{c}{Energy} \\ \midrule
 \multirow{3}{*}{TimeBridge}  
 & $\beta_{min}$ & \{0.1\}   & 0.1 & 0.1  & 0.1 \\
  & $\beta_{d}$ & \{0.2, 0.5, 1\} &  0.2 & 0.2  & 0.2  \\
  & $\sigma_{data}$ & \{0.05, 0.1, 0.5\} & 0.5 & 0.1 & 0.05$^\dagger$
\\ \bottomrule
\end{tabular}%
}
\end{table}

\paragraph{\textbf{Details for sampling from GP Prior}}

We sample $\rvx_T$ in \Eqref{eq:gp_prior} as follows:
\begin{equation}
\label{eq:gp_exp}
\rvx_T \sim \mathcal{N}(\boldsymbol{m}(\Omega), \boldsymbol{\Sigma}(\Omega)), \quad \Omega = \{1,\ldots,\tau\}.
\end{equation}
Specifically, we set $\boldsymbol{m}(\Omega)$  and $\boldsymbol{\Sigma}(\Omega)$ by $\boldsymbol{m}(\Omega) = \boldsymbol{\mu}$ and $\boldsymbol{\Sigma}(\Omega)_{i,j} = \texttt{diag}
(\boldsymbol\sigma^2)_{i,j} + \eta\exp\left(-|i - j|^2\right)$ for timestamps $i$ and $j$ in \Eqref{eq:mean}.

\paragraph{\textbf{Hyperparameter Setups}}

For experimental setups, we follow the base setups of \citet{yuan2024diffusionts}. 
Therefore, we try to use the minimal search space for determining only for diffusion scheduler $\rvx_t = \alpha_t \rvx_0 + \sigma_t \rvvarepsilon$, where $\alpha_t$ and $\sigma_t$ are the functions of times. 

As demonstrated in EDM \cite{karras2022elucidating}, we can adjust $\alpha_t$ and $\sigma_t$ based on $\beta_{min}$ and $\beta_d$ by reparameterizing $\alpha_t$ and $\sigma_t$ using $\beta_{min}$ and $\beta_{d}$ as follows:
\begin{align}
    \alpha_t &= e^{-\frac{1}{4}\beta_d t^2 - \frac{1}{2} \beta_{min} t}\\ 
    \sigma_t &= e^{\frac{1}{2} \beta_d t^2 + \beta_{min} t}-1.
\end{align} 

The summarized hyperparameters used for unconditional generation and conditional generation are demonstrated in Tables \ref{tab:uncond-setting} and \ref{tab:cond-setting},  respectively. In the GP setups, we additionally search the kernel parameter $\eta$. 

We unified the sampling settings with the second-order Heun sampler with 40 steps, which has 119 NFE. We use the churn ratio $s=0.33$ for stochastic sampling in Algorithms \ref{alg:hybrid_sampler}.


\end{document}